\documentclass[lettersize,journal]{IEEEtran}
\usepackage{amsmath,amsfonts}
\usepackage{algorithmic}
\usepackage{algorithm}
\usepackage{array}
\usepackage[caption=false,font=normalsize]{subfig}

\usepackage{float} 
\usepackage{textcomp}
\usepackage{stfloats}
\usepackage{url}
\usepackage{verbatim}
\usepackage{graphicx}
\usepackage{cite}
\usepackage{color}
\usepackage{colortbl}

\def\datasetName{UIIS10K}
\def\methodName{UWSAM}
\def\eg{\textit{e.g.}}
\def\ie{\textit{i.e.}}

\begin{document}

\title{Advancing Marine Research: UWSAM Framework and UIIS10K Dataset for Precise Underwater Instance Segmentation}

\author{Hua Li$^1$,\,Shijie Lian$^1$,\,Zhiyuan Li,\,Runmin Cong, \IEEEmembership{Senior Member, IEEE},\,Chongyi Li \IEEEmembership{Senior Member, IEEE}\\ \,Laurence T. Yang \IEEEmembership{Fellow, IEEE},\,Weidong Zhang, \IEEEmembership{Senior Member, IEEE}, Sam Kwong, \IEEEmembership{Fellow, IEEE},
\thanks{$^1$ These authors contributed equally to this work.}
\thanks{Hua Li is with the School of Computer Science and Technology, Hainan University, Haikou, 570100 China.}
\thanks{Shijie Lian is with the School of Computer Science and Technology, Huazhong University of Science and Technology, Wuhan, 430074 China, and also with the Zhongguancun Academy, Beijing, 100080, China.}
\thanks{Zhiyuan Li is with the School of Computer Science and Technology, Hainan University, Haikou, 570100 China.}
\thanks{Runmin Cong is with the School of Control Science and Engineering, Shandong University, Jinan, 250100 China.}
\thanks{Chongyi Li is with the College of Computer Science, Nankai University, Tianjin, 300071 China.}
\thanks{Laurence T. Yang is with the School of Computer Science and Artificial Intelligence, Zhengzhou University, Zhengzhou, China, and also with the School of Computer Science and Technology, Huazhong University of Science and Technology, Wuhan, 430074 China.}
\thanks{Weidong Zhang is with the Department of Automation, Shanghai Jiao Tong University, Shanghai 200240, China.}
\thanks{Sam Kwong is with the School of Data Science, Lingnan University, Hong Kong, China.}
}



\maketitle

\begin{abstract}

With recent breakthroughs in large-scale models, the Segment Anything Model (SAM) has demonstrated significant potential in a variety of visual applications.
However, due to the lack of underwater domain expertise, SAM and its variants face performance limitations in end-to-end underwater instance segmentation tasks, while their higher computational requirements further hinder their application in underwater scenarios.
To address this challenge, we propose a large-scale underwater instance segmentation dataset, \datasetName, which includes 10,048 images with pixel-level annotations for 10 categories.
Then, we introduce \methodName, an approach designed for automatic and accurate segmentation of underwater instances.
UWSAM efficiently distills knowledge from the SAM ViT-Huge image encoder into the smaller ViT-Small image encoder via the Mask GAT-based Underwater Knowledge Distillation (MG-UKD) method for effective visual representation learning.
Furthermore, we design an End-to-end Underwater Prompt Generator (EUPG) for UWSAM, which automatically generates underwater prompts instead of explicitly providing foreground points or boxes as prompts, thus enabling the network to locate underwater instances accurately for efficient segmentation.
Comprehensive experimental results indicate that UWSAM-Student achieves higher segmentation accuracy than WaterMask by 1.3, 0.5, and 1.1 mAP$^s$ on UIIS10K, UIIS, and USIS10K respectively, while UWSAM-Teacher surpasses the similarly sized USIS-SAM by 4.8, 1.8, and 2.5 mAP$^s$ on the same datasets.
Datasets and codes are available at \url{https://github.com/LiamLian0727/UIIS10K}.
\end{abstract}


\section{Introduction}

\begin{figure}[!t]
  \centering
  \subfloat[]{
    \includegraphics[width=\linewidth]{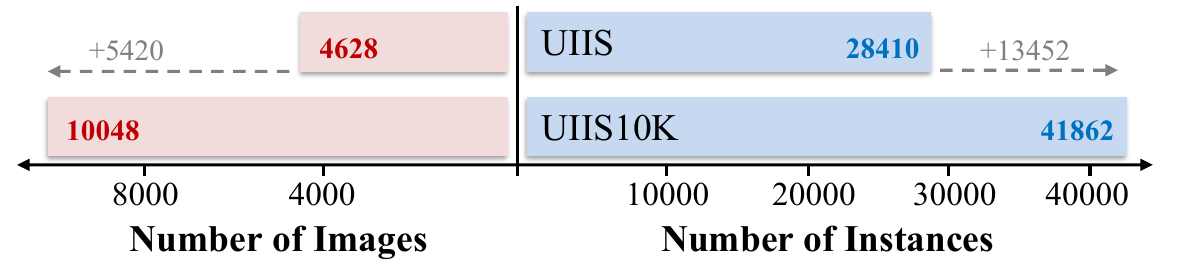}
    \label{fig:showa}
  }\\[-0ex]
  \subfloat[]{
    \includegraphics[width=\linewidth]{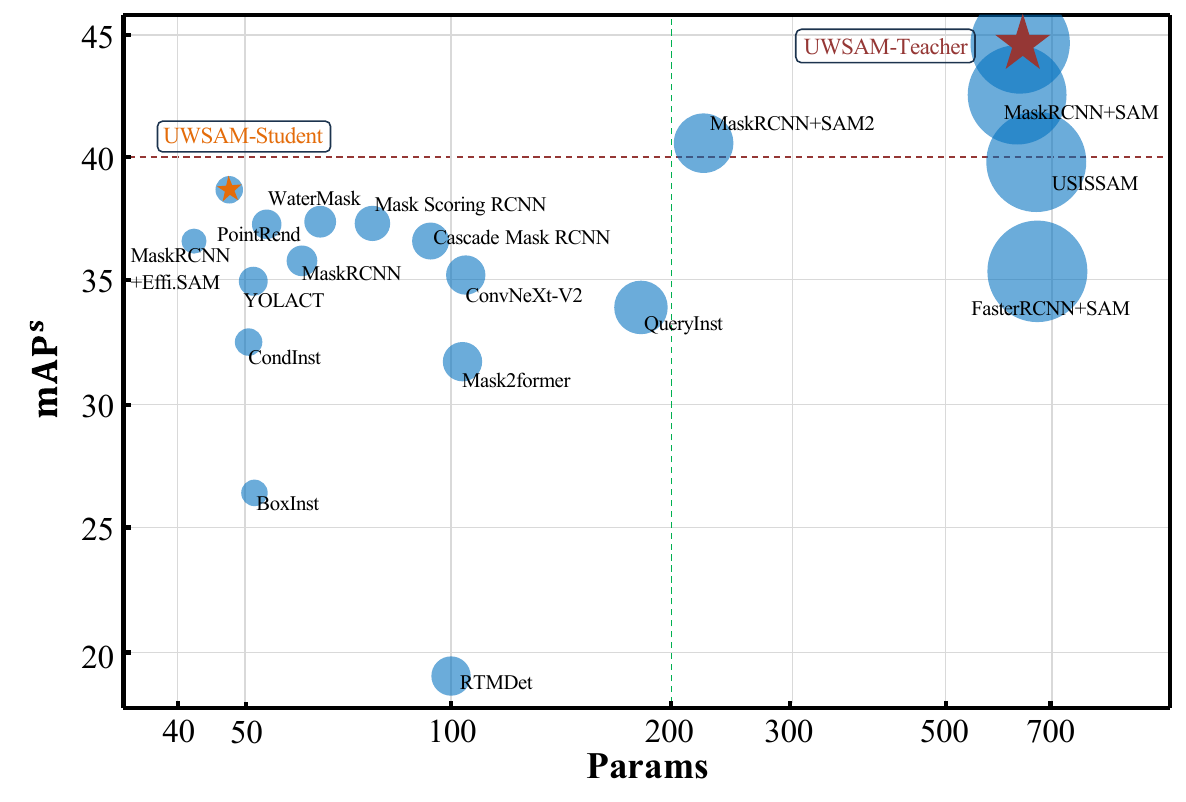}
    \label{fig:showb}
  }
  \caption{(a) Comparison of the UIIS and UIIS10K datasets in terms of the number of images and instances. UIIS10K significantly expands the dataset scale. (b) Comparison of models in terms of parameters versus mAPs, UWSAM-Teacher and UWSAM-Student achieving strong performance at different scales. Effi. SAM denotes MaskRCNN\cite{MaskRCNN_2017_ICCV} + SAM\cite{EfficientSAM_2024_CVPR}. mAP$^s$ is mask AP metrics}
  \label{fig:show}
\end{figure}

Underwater visual understanding and dense prediction tasks have recently received increasing attention in the computer vision field in terms of exploring and protecting underwater ecosystems \cite{CoralSCOP_2024_CVPR, WaterMask_2023_ICCV, MarineInst_2024_ECCV, USISSAM_2024_ICML, USOD10K_2023_TIP}.
Instance segmentation, a fundamental and crucial vision task, aims to segment all objects in an image and distinguish them from each other, thereby enabling detailed and accurate scene understanding \cite{MaskRCNN_2017_ICCV}.
By providing pixel-level object localization, instance segmentation provides the foundation for finer instance-level perception and a wide range of applications \cite{SOLO_2022_TPAMI}.
Underwater Images Instance Segmentation (UIIS) shows great potential for expanding underwater vision applications, such as discovery of marine relics, exploration of marine resources, underwater human-computer interaction, and comprehensive analysis of underwater images \cite{WaterMask_2023_ICCV}.
The UIIS dataset \cite{WaterMask_2023_ICCV} is the first large-scale dataset for underwater instance segmentation that provides pixel-level annotations in different scenes and object classes, and proposes WaterMask, a detail-enhanced boundary-aware framework for underwater visual features, which provides a solid foundation for subsequent research.
However, the UIIS dataset contains only 4628 images, limiting its scale for large-model training. In addition, as a CNN-based approach, WaterMask may achieve lower segmentation accuracy than more recent transformer-based foundation segmentation models such as SAM \cite{SAM_2023_ICCV}.

\IEEEpubidadjcol

In recent years, large-scale language models (LLMs) such as the Generative Pretraining Transformer (GPT)-4 \cite{GPT4_2023_arXiv}, Language Learning for Adaptive Multitasking Architecture (LLaMA) \cite{Llama_2023_arXiv}~and Pathways Language Model (PaLM) \cite{PaLM_2023_JMLR}~have sparked a revolution in the field of natural language processing (NLP).
These foundational models exhibit excellent generalization capabilities and perform well in numerous open-world language tasks. 
Inspired by the success of LLMs, visual base models such as Contrastive Language-Image Pre-Training model (CLIP) \cite{CLIP_2021_PLMR}, Segment Anything Model (SAM) \cite{SAM_2023_ICCV}~ and Segment Anything Model 2 (SAM2) \cite{SAM2_2024_arXiv}~also emerged. 
The introduction of these foundation models continues to drive researchers' exploration in the field of computer vision.
Notably, SAM and SAM 2 exhibit robust performance in various segmentation tasks with their powerful encoder-decoder transformer framework and large-scale datasets.
With fine-tuning or appropriate modifications, it has significant potential in the field of marine science.
Despite these advantages, deploying models like SAM on edge computing devices such as unmanned underwater vehicles still suffers from performance bottlenecks.
The primary issue stems from the complexity of the model architectures, especially the image encoders (e.g., ViT-H \cite{ViT_2020_ICLR}~and Hiera-L \cite{PaLM_2023_JMLR}), which contain 636M and 224M parameters, respectively.
Consequently, directly utilizing these models for instance segmentation requires considerable computational and memory costs, greatly limiting their applicability in underwater environments.


To address this challenge, several recent studies have proposed strategies to avoid incurring huge costs in prompt-based instance segmentation. 
For example, FastSAM\cite{FastSAM_2023_arXiv}~utilizes a real-time CNN-based architecture to reduce computational overhead, while EfficientSAM \cite{EfficientSAM_2024_CVPR}~leverages the well-known Masked Autoencoders (MAE) method \cite{MAE_2022_CVPR}~and the SAM model to refine a lightweight and efficient image encoder.
Despite these efforts, obtaining strong performance with SAM and its variants in underwater instance segmentation typically requires substantial downstream data for post-training \cite{CoralSCOP_2024_CVPR, MarineInst_2024_ECCV}. For example, in underwater salient instance segmentation, the USIS-SAM \cite{USISSAM_2024_ICML} approach utilizes the USIS10K dataset and an adapter-based fine-tuning strategy to achieve competitive results. However, since USIS10K is specifically designed for salient segmentation, it does not provide annotations for all instances present in the scene. To address this limitation and support large-scale post-training and evaluation of models such as SAM for underwater instance segmentation, we introduce the \datasetName~dataset, containing 10,048 underwater images with pixel-level annotations in 10 categories such as fish, coral, ruins, and human. This dataset is intended to support instance segmentation tasks and facilitate the development of models better adapted to underwater environments.
As shown in Figure \ref{fig:show}(a), the UIIS10K dataset contains 5,420 additional images and 13,452 more pixel-level instance masks than the UIIS dataset, offering a larger resource for training and evaluation.

Simultaneously, we introduce UWSAM, a multi-objective underwater image instance segmentation method designed for the intrinsic features of underwater images with the SAM framework. UWSAM uses the Mask GAT-based Underwater Knowledge Distillation (MG-UKD) algorithm to train ViT-Small backbone \cite{ViT_2020_ICLR}~as the image encoder, while the End-to-End Underwater Prompt Generator (EUPG) combined with the SAM Mask Decoder guides the network to achieve end-to-end segmentation.
MG-UKD capitalizes on a fundamental characteristic of underwater imagery: underwater instances (e.g., schools of fish, coral reefs) often form groups and clusters, leading to distinct regions of an underwater image sharing visual similarities.
Leveraging this prior knowledge, we randomly mask the features output by the ViT-Small image encoder and employ a Graph Attention Network (GAT) to reconstruct these features to align with those extracted by the SAM ViT-Huge image encoder. 
By distilling knowledge from a larger, more complex model to a smaller, more efficient one, the resulting model becomes better suited for underwater deployment. 
By incorporating underwater domain-specific knowledge and employing effective feature reconstruction, our approach successfully addresses the unique challenges of underwater environments, including variable lighting conditions, turbidity, and the presence of diverse marine species.

Furthermore, unlike traditional methods that depend on external object detectors to generate point or bounding box (BBox) prompts for SAMs, subsequently encoded into prompt features by a SAM prompt encoder, EUPG directly integrates the generation of underwater prompt features within the model itself, enabling end-to-end training and inference. 
By incorporating EUPG, our approach eliminates the need for manually provided prompts or external detectors, significantly enhancing the efficiency and adaptability of underwater segmentation. 
Experimental results indicate that EUPG not only improves segmentation accuracy but also reduces computational complexity, making it highly suitable for deployment on resource-constrained underwater platforms.

To evaluate the effectiveness of our proposed methods, we conducted extensive experiments on the \datasetName~datasets, USIS10K datasets \cite{USISSAM_2024_ICML}, and UIIS datasets \cite{WaterMask_2023_ICCV}, comparing our approach with state-of-the-art SAM-based methods, as well as other underwater and general instance segmentation methods. 
The results demonstrate that our approach significantly outperforms these existing methods in terms of segmentation accuracy, robustness, and efficiency. 
Our methods particularly excel in addressing the challenges unique to underwater environments, such as variable lighting, occlusions, and the complex appearance of underwater objects.
As shown in Figure~\ref{fig:show}(b), our UWSAM-Teacher achieves the highest performance on the UIIS10K dataset, with an mAP$^{\text{s}}$ of 44.6, outperforming USIS-SAM 6.8 APs. In addition, UWSAM-Student achieves an mAP$^{\text{s}}$ of 38.7, exceeding WaterMask by 1.3 APs, demonstrating the effectiveness of our approach across different model sizes.

The main contributions are concluded as follows:
\begin{itemize}
\item We present \datasetName, a large underwater instance segmentation dataset containing 10,048 images with pixel-level annotations for 10 categories with \methodName~model. As far as we know, this is the largest underwater instance segmentation dataset available and can be used as a benchmark for evaluating underwater segmentation methods.
\item We propose the Mask GAT-based Underwater
Knowledge Distillation (MG-UKD) algorithm for \methodName, which distills knowledge from large SAM ViT-Huge encoders into smaller ViT-Small encoders, optimized for underwater environments. MG-UKD reduces computational complexity and improves segmentation accuracy under challenging conditions such as low visibility and varying lighting, making it well-suited for real-world underwater applications.
\item We developed the End-to-End Underwater Prompt Generator (EUPG) for \methodName, which directly generates prompts with positional information and contextual details, eliminating the need for external detectors or manual inputs, and enabling efficient end-to-end underwater segmentation.
\item Extensive public evaluation metrics and numerous experiments confirm the effectiveness of our \datasetName~dataset and \methodName~model.
\end{itemize}

\section{RELATED WORK}

\subsection{Underwater Instance Segmentation Dataset}


Unlike terrestrial segmentation models, underwater instance segmentation faces unique challenges due to optical distortions that occur underwater, such as light scattering, refraction, and color attenuation  \cite{Akkaynak_2017_CVPR}.
These factors result in images that are often unclear, poor contrast, blurry, or with color distortion, posing difficulties for general segmentation models. 
This degradation can be quantitatively described by the SeaThru model \cite{SeaThru_2019_CVPR}:
\begin{equation}
\label{eq:seathru_mdoel}
I_c = J_c\cdot exp(-\beta_c^D\cdot z) + B^\infty_c\cdot (1-exp(-\beta^B_c \cdot z))\text{,}
\end{equation}
where channel $c \in \left \{ R, G, B\right \}$, $I_c$~is the image captured by the camera, $J_c$~is the scene radiance, $B^\infty_c$~is the veiling light, $\beta_c^D$~and $\beta^B_c$~controls are the light attenuation coefficient and backscattering coefficient of the image in RGB channel \cite{Akkaynak_2018_CVPR, DeepSeeColor_2023_ICRA}, and $z$~is the scene depth.
In addition to quality degradation, marine organisms also display significant diversity in shape, size, and color \cite{UnderwaterOD1_2020_ECCV, UnderwaterOD2_2022_JOE}, all of which result in general models trained on terrestrial instance segmentation datasets not being able to achieve satisfactory results on underwater instance segmentation tasks.
Therefore, the community has been focusing on annotating underwater segmentation datasets in recent years.
The TrashCan dataset \cite{TrashCan_2020_arXiv}~focuses on marine environmental protection and provides annotated underwater trash images for training autonomous underwater vehicles (AUVs) detectors.
The DeepFish dataset \cite{DeepFish_2022_IJCNN}~contains in-air fish images from wholesale fish markets for fish species classification and size estimation.
The UIIS dataset \cite{WaterMask_2023_ICCV}~is the first generalized underwater instance segmentation dataset containing 4628 images in 7 categories with pixel-level annotations, thus encouraging the development of more accurate and context-sensitive segmentation solutions.
The USIS10K dataset \cite{USISSAM_2024_ICML}~focuses on another variant of instance segmentation, underwater salient instance segmentation, and is designed to help segmentation models focus on segmenting valuable underwater salient objects.
However, the above datasets are either designed for a particular task, or the number of annotations is insufficient. Since large-scale benchmark datasets play an essential role in developing underwater instance segmentation methods in the large model era, we constructed a new challenging dataset, called USIS10K, which covers 10,048 underwater images with fine-grained annotations for 10 categories of underwater instances.
Compared to other existing datasets, our dataset is larger and has more diverse categories of objects.
The refUDS dataset \cite{refUDS_2025_IF}~extends UIIS and USIS10K by pairing detailed textual descriptions of underwater images with their corresponding segmentation masks. This resource enables the development and training of models capable of jointly processing visual and linguistic inputs.

\subsection{Instance Segmentation Model}

Terrestrial Instance Segmentation research has made remarkable progress over the past decade\cite{ISSurvey_2020_IJMIR}.
Classical instance segmentation methods typically rely on the two-stage framework of Mask R-CNN \cite{MaskRCNN_2017_ICCV}, which uses Region Proposal Networks (RPN) \cite{FastRCNN_2015_ICCV}~to generate bounding boxes and RoIAlign \cite{MaskRCNN_2017_ICCV}~to extract instance features from the Feature Pyramid \cite{FPN_2017_CVPR}~for pixel-level mask prediction. 
With the use of Transformer \cite{Transformer_2017_NIPS}~in computing, QueryInst \cite{QueryInst_2021_ICCV}, inspired by DETR \cite{DETR_2020_ECCV}, treats the object of interest as a learnable segmentation query and performs the segmentation using dynamic mask heads, which outperforms previous techniques in accuracy and speed.
In addition, Mask2Former \cite{Mask2Former_2022_CVPR}~improves segmentation performance by limiting the cross-attention range in the transformer decoder, enabling more effective extraction of local features.
Although the history of underwater instance segmentation is short compared with instance segmentation tasks, a considerable number of research works have been related to this topic.
WaterMask \cite{WaterMask_2023_ICCV}~uses multi-scale refinement and graph attention mechanisms to combat underwater visual artifacts and thus improve underwater instance segmentation accuracy.
TC-USOD \cite{USOD10K_2023_TIP}~deeply fuses RGB images with depth images in the transformer encoder and uses a lightweight convolutional decoder to segment the salience mask of an object.
Recently, foundation models (\eg, CLIP\cite{CLIP_2021_PLMR}, GLIP \cite{GLIP_2022_CVPR}, ALIGN \cite{ALIGN_2021_PMLR}, SAM\cite{SAM_2023_ICCV}, SAM2 \cite{SAM2_2024_arXiv}) have received a lot of attention in the computer vision community. 
Among them, SAM \cite{SAM_2023_ICCV}, an interactive segmentation model trained on the large SA-1B dataset, receives various user inputs (\eg, points, boxes, and masks) in a semantically-agnostic manner to accurately segment visual objects, exhibiting robust zero-point generalization and promptable segmentation performance.
CoralSCOP\cite{CoralSCOP_2024_CVPR}~applies SAM to the coral reef segmentation task, solving the coral reef classification challenge by adding a parallel semantic branch to the SAM decoder and introducing negative masks with “non-coral” labels during training to help the network converge.
Furthermore, MarineInst \cite{MarineInst_2024_ECCV}~combines SAM with frozen VLMs such as CLIP \cite{CLIP_2021_PLMR}~or MarineGPT \cite{MarineGPT_2023_arXiv}, to analyze underwater imagery through instance-level visual descriptions that can support a wide range of marine visual analysis and scene understanding tasks.
Nonetheless, all of the above models rely on external explicit prompts from user inputs and cannot accomplish end-to-end underwater instance segmentation.
In addition, the large number of parameters further limits their application in scenarios such as underwater vehicles.
To this end, we propose the MG-UKD algorithm to reduce the number of parameters in \methodName~through knowledge distillation and introduce the EUPG module to generate visual prompts for automatic underwater instance segmentation.

\begin{figure*}[!t]
\centering
\includegraphics[width=1\linewidth]{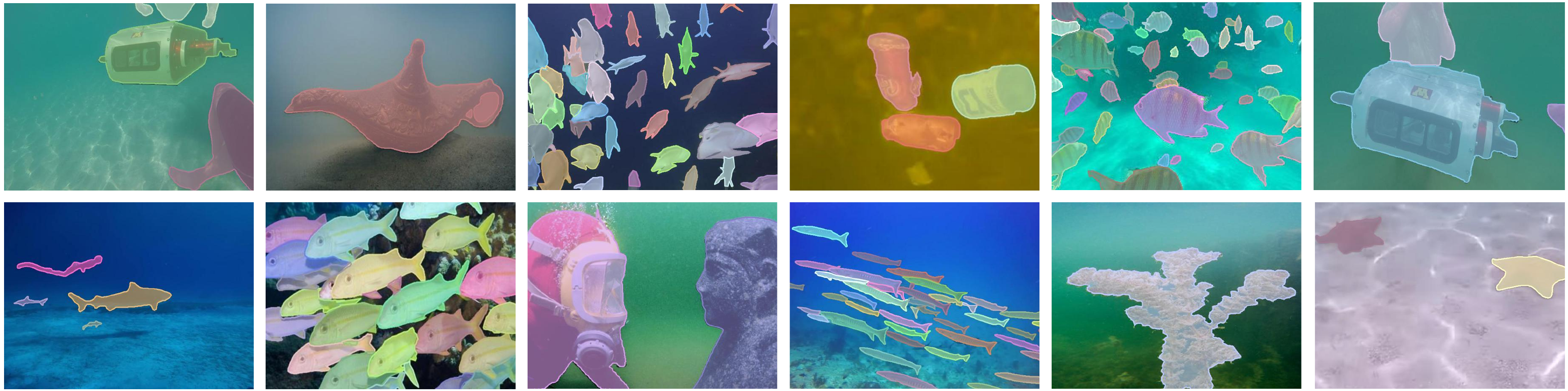}%
\caption{The figure shows pixel-level instance segmentation in \datasetName~dataset of specific object classes: fish, reptiles, arthropods, corals, mollusks, plants, ruins, garbage, human, and robots, and the segmented masks are superimposed on the image.}
\label{fig:dataset_show}
\vspace{-2mm}
\end{figure*}

\subsection{Knowledge Distillation}
Knowledge Distillation (KD) has become a widely-used technique for transferring knowledge from a large teacher model to a smaller student model while maintaining high performance \cite{KDReview_2022_PMAI}. 
The foundational work by Hinton et al. \cite{KD_2014_NIPS}~introduced the idea of using soft labels from a teacher to guide the student model, improving its ability to generalize despite having fewer parameters.
Since then, numerous studies have explored various aspects of KD, aiming to enhance its efficiency and effectiveness. 
MGD \cite{MGD_2022_ECCV}~masks random pixels from the student's features and allows the student to reconstruct the teacher's complete features through a simple convolutional operation, thus encouraging the student to focus on the important features of the data.
DMAE \cite{DMAE_2023_CVPR}~focuses on extracting knowledge from pre-trained models such as Masked Autoencoders (MAE) \cite{MAE_2022_CVPR}, and performing efficient knowledge distillation by minimizing the distance between the intermediate features extracted by the teacher and the students on the visible input patches.
EfficientSAM \cite{EfficientSAM_2024_CVPR}~also applies MAE for knowledge distillation to the SAM image encoder, introducing the SAMI approach. 
However, unlike DMAE, SAMI masks only the student model's features and uses an additional cross-attention decoder to reconstruct these features to be consistent with the teacher model.
However, since backbone models are generally pre-trained on terrestrial datasets and lack underwater visual knowledge, directly using their distilled results for underwater instance segmentation may lead to sub-optimal performance.

\section{\datasetName~DATASET}

In order to advance the progress of UIIS research, we collected and annotated a new large-scale benchmark dataset named \datasetName, which contains a total of 10,048 RGB underwater images with pixel-level instance annotations.
Some samples of this dataset are shown in Fig. \ref{fig:dataset_show}.
In this section, we will describe the construction process of \datasetName~in detail and analyze its characteristics comprehensively.

\begin{table}[!t]
    \caption{Comparison with the existing underwater instances dataset. Numbers denotes the number of images in each dataset}\vspace{-3mm}
    \begin{center}
    \renewcommand{\arraystretch}{1.1}
    \setlength{\tabcolsep}{3.9mm}
    {\begin{tabular}{c|ccc}
    \hline\hline
    \rowcolor[RGB]{200,200,200}\textbf{Dataset} & \textbf{UIIS} \cite{WaterMask_2023_ICCV} & \textbf{TrashCan} \cite{TrashCan_2020_arXiv} &  \textbf{\datasetName~(Ours)}\\
    \hline
    Years & 2023 & 2020 & 2025\\
    Numbers& 4,627  & 7,212 & \textbf{10,048} \\\hline\hline
    \end{tabular}}
    \end{center}
    \label{tab:is.comp}
    \vspace{-2mm}
\end{table}

\begin{table}[t]
    \caption{Category descriptions of \datasetName~dataset.}\vspace{-3mm}
    \renewcommand{\arraystretch}{1.1}
    \begin{center}
    \renewcommand{\arraystretch}{1.1}
    {\begin{tabular}{c|c}\hline\hline
        \rowcolor[RGB]{200,200,200}\textbf{Category} & \textbf{Descriptions}\\\hline
        Fish & Fish and other marine vertebrates \\\hline
        Reptiles & Marine reptiles, e.g. turtles, sea snakes, etc. \\\hline
        Artiodactyla & Shelled marine animals like shrimps, crabs, sea spider \\\hline
        Mollusk & Underwater mollusks, e.g. shells, octopuses, starfish, etc. \\\hline
        Corals & Corals and coral reefs \\\hline
        Plants & Aquatic plant \\\hline
        Garbage & Household trash, industrial waste, broken machine, etc. \\\hline
        Ruins & Underwater wrecks and ruins, e.g. shipwrecks, etc. \\\hline
        Robots & AUV, ROV and other underwater robots \\\hline
        Human & Human divers (including body diving equipment) \\
        \hline\hline
    \end{tabular}}
    \end{center}
    \label{tab:label}
    \vspace{-5mm}
\end{table}

\subsection{Dataset Collection \& Annotation \& Splitting}\label{subsec:dataset}

The construction of the \datasetName~dataset is divided into the following three steps: 1) image collection and filtering, 2) image annotation, and 3) dataset splitting.

\noindent\textbf{Dataset Collection.} In order to enrich the variety of images in the dataset as much as possible to cover more underwater scenarios and lighting conditions.
We collected about 40,000 images from the Internet and open-source underwater datasets from various domains, such as underwater image enhancement \cite{UIEB_2019_TIP, UnderwaterHL_2020_TPAMI, LSUI_2023_TIP}, underwater semantic segmentation \cite{SUIM_2020_IROS}, underwater instance segmentation \cite{WaterMask_2023_ICCV, TrashCan_2020_arXiv}, and underwater salience detection \cite{UFO120_2020_Robotics, USOD10K_2023_TIP, USISSAM_2024_ICML}.
These images come from various underwater environments, including the deep sea, shallow waters, and lakes, covering tasks such as marine resource exploration, intelligent human-machine collaboration, and underwater environmental conservation.
Afterwards, we had two volunteers filter the candidate images to remove repetitive, damaged, or non-underwater environments images. Finally, we carefully annotated the remaining 14,500 images.

\noindent\textbf{Dataset Annotation.}
Leveraging SAM's strong zero-shot generalization ability, which significantly reduces the labeling workload, we utilize SAM\cite{SAM_2023_ICCV}~and EfficientSAM\cite{EfficientSAM_2024_CVPR}~models for assisted labeling. 
After the models generate the initial labels, volunteers manually refine them and assign category labels to each mask.
We recruited 16 volunteers to label the dataset. Before starting the annotation process, they received training in the following four areas:
(1) Classification of common organisms in underwater scenes.
(2) The previous dataset construction process and basic image labeling methods.
(3) How to effectively input prompts for semi-automatic labeling using the SAM and EfficientSAM models.
(4) How to manually refine the masks after semi-automatic annotation.

\begin{figure}[!t]
\centering
\includegraphics[width=1\linewidth]{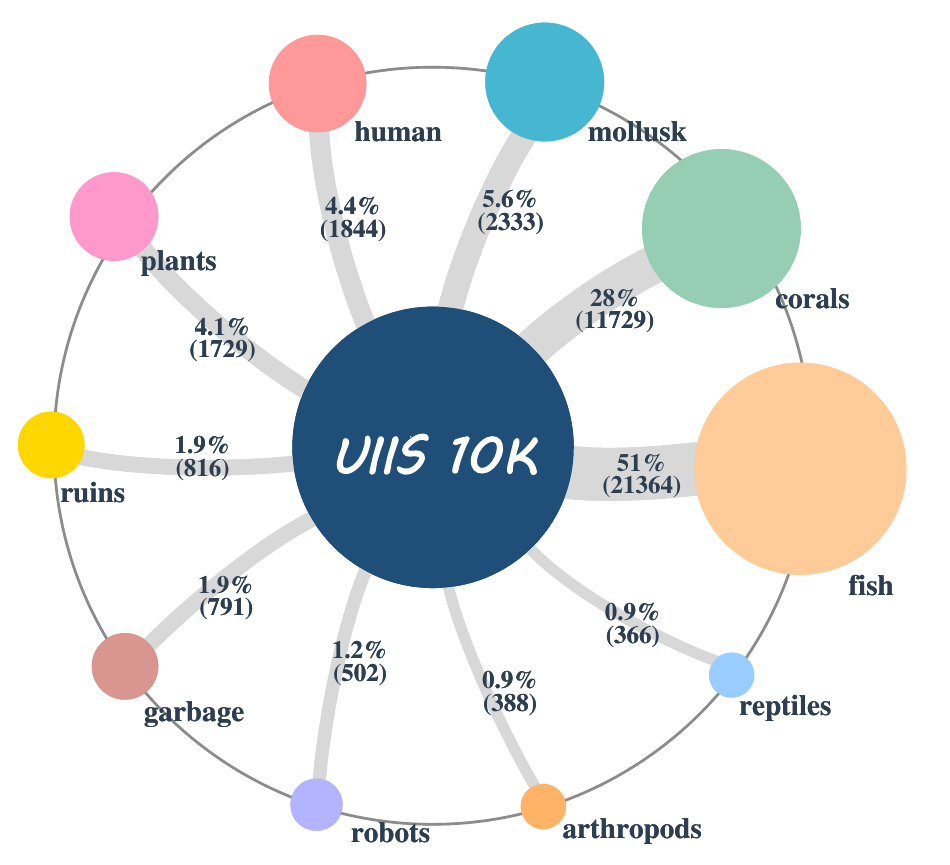}%
\vspace{-2mm}
\caption{The number of instances per category in the \datasetName~dataset. There are 41,862 carefully labeled instance masks in the UIIS10K dataset.
\vspace{-2mm}
}
\label{fig:uiis_num}
\end{figure}

We used sparse annotation polygons to label each instance in the dataset. The annotated data will be stored in the widely-used COCO-style format \cite{MSCOCO_2014_ECCV}, ensuring compatibility with most popular frameworks and models.
Each image will be labeled by at least two volunteers and reviewed by a third volunteer.
For images collected from the underwater instance segmentation dataset, two volunteers will manually refine the original instance annotations and reassign labels.
A third volunteer will then select, refine, and merge their annotations, as well as check whether the image consists of unlabeled instances.
For images from the underwater salient segmentation dataset, in addition to refining and reassigning labels for the existing masks, the two volunteers will use SAM and EfficientSAM to annotate non-salient instances that were previously unmarked. The third volunteer will choose the best annotations and refine them.
For other images, two volunteers will use SAM and EfficientSAM to annotate all objects of interest at the instance level. The third volunteer will select and merge their annotations, further refining the masks and category labels to ensure accuracy and precision.
When using the model for assisted annotation, SAM and EfficientSAM each generate three candidate masks for every set of prompts provided by the volunteer. The volunteer then selects the most suitable mask and manually refines it.

To improve the accuracy of labeling categories, we classified potentially ambiguous objects following the guidelines outlined in \cite{conf_2001_ceana, conf_2009_ETI}. Additionally, we excluded images containing instances where consensus on the annotation could not be reached.
Finally, we obtained 10,048 images that make up the \datasetName~dataset.

\noindent\textbf{Dataset Splitting.} 
To ensure practical training and reliable test results for deep learning methods on the \datasetName~dataset, it is essential to have sufficient samples of each category in both the training and test sets. Therefore, we follow the \datasetName~split of approximately 8:2 for the training and test sets. Specifically, the \datasetName~dataset is divided into 8,083 samples for training and 2,010 samples for validation and testing.

\begin{figure}[!t]
\centering
\includegraphics[width=1\linewidth]{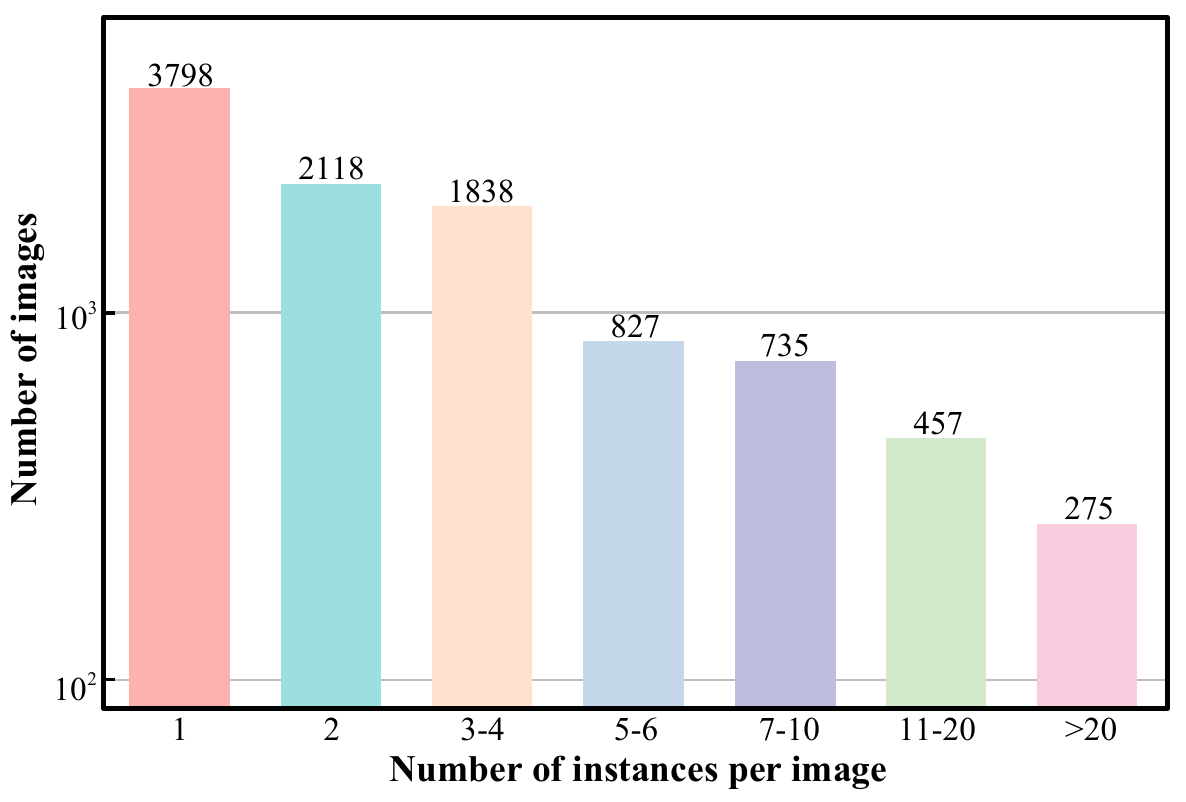}%
\vspace{-2mm}
\caption{Distribution of the number of instances per image in the \datasetName~dataset. 
\vspace{-2mm}
}
\label{fig:num_ins_per_img}
\end{figure}

\subsection{Dataset Characteristics and Statistics}
In this subsection, we illustrate the basic information, characteristics, and challenges of the \datasetName~dataset.

\noindent\textbf{Number and Category of Dataset.}
A large number of underwater images from different scenes and containing different categories is essential to improve the generalization ability and avoid overfitting of the network in complex marine environments.
To address this, we extended the UIIS dataset \cite{WaterMask_2023_ICCV}~to provide detailed annotations for 10,048 images from different scenes.
As shown in Table \ref{tab:is.comp}, \datasetName~is the largest existing instance segmentation dataset for underwater scenarios.
Furthermore, in order to make the \datasetName~dataset applicable to a broader range of downstream tasks, we labeled the instances in \datasetName~according to the 10 categories shown in Table \ref{tab:label}.
The \datasetName~dataset includes categories such as fish, reptiles, artiodactyla, and ollusks, which are the main research components of marine ecological exploration.
It also provides pixel-level annotations for corals, plants, and garbage, which are primary targets in marine ecological conservation.
In Fig. \ref{fig:uiis_num}, we also count the number of masks for each category. 
the \datasetName~dataset contains a total of 41,862 masks, with the fish and coral categories having the highest number of masks, which are the two most common objects in underwater environments.
In addition, the dataset contains annotations for human divers, robots, and ruin, which are valuable for training applications involving human-robot-object intelligent cooperation.
Detailed category definitions are given in Table \ref{tab:label}.
In addition, the \datasetName~dataset divides all annotations into three parts: category labels, instance masks, and bounding boxes. This provides the possibility of using the \datasetName~dataset for other downstream tasks in underwater scenarios such as object detection and semantic segmentation.

\noindent\textbf{Number of instances in the image.} In the \datasetName~dataset, a single image often contains multiple instances. As shown in Fig. \ref{fig:dataset_show}, most of these images typically have more than one instance. In Fig. \ref{fig:num_ins_per_img}, we provide statistics on the number of instances in the dataset. Specifically, for the \datasetName~dataset, 22.83\% of the scenes have more than 5 instances, 8.45\% of the scenes have more than 10 instances, and the image with the most instances contains 81 instances.  The increase in the number of instances in underwater images is often due to factors such as fish groups or coral clusters. For example, in the image in the second row, second column of Fig. \ref{fig:dataset_show}, instances tend to cluster closely together and occlude each other, which presents a greater challenge for accurately segmenting the instance boundaries.

\begin{figure}[!t]
\centering
\includegraphics[width=1\linewidth]{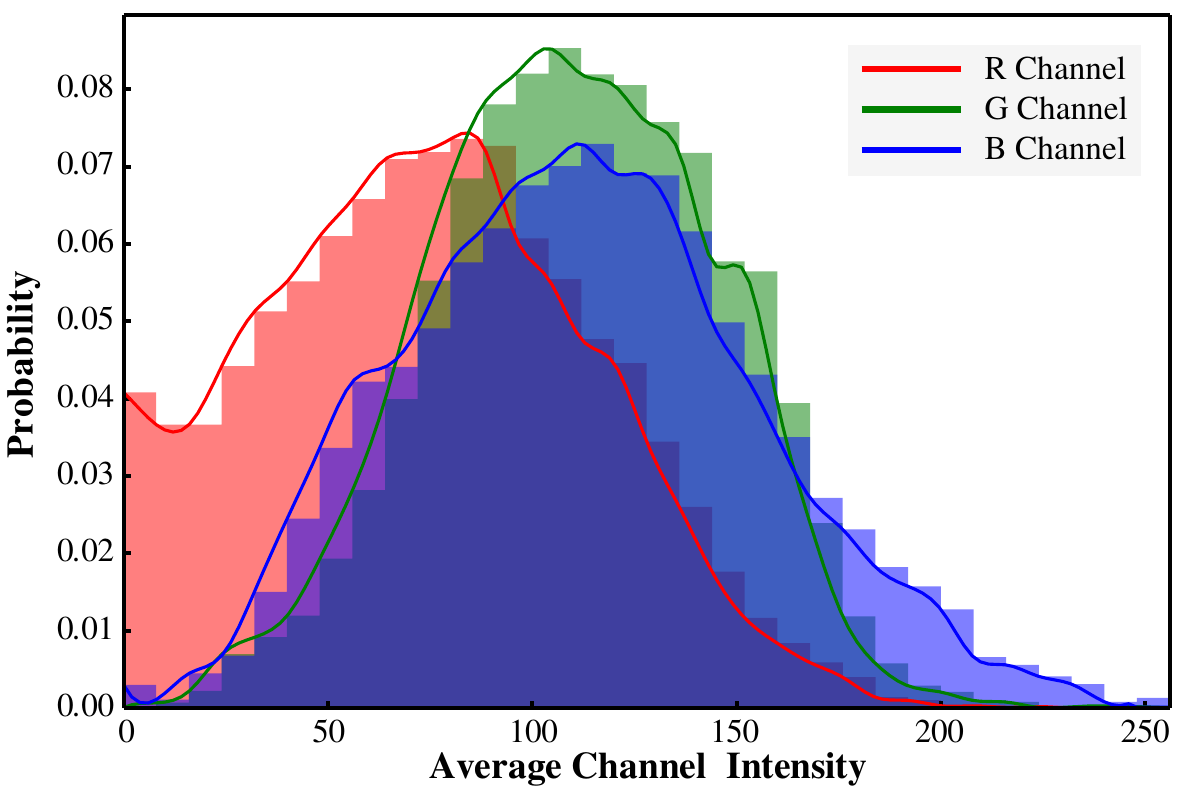}%
\vspace{-2mm}
\caption{Average channel intensity in \datasetName~with proportion.}
\vspace{-2mm}
\label{fig:channel}
\end{figure}

\begin{figure}[!t]
\centering
\includegraphics[width=1\linewidth]{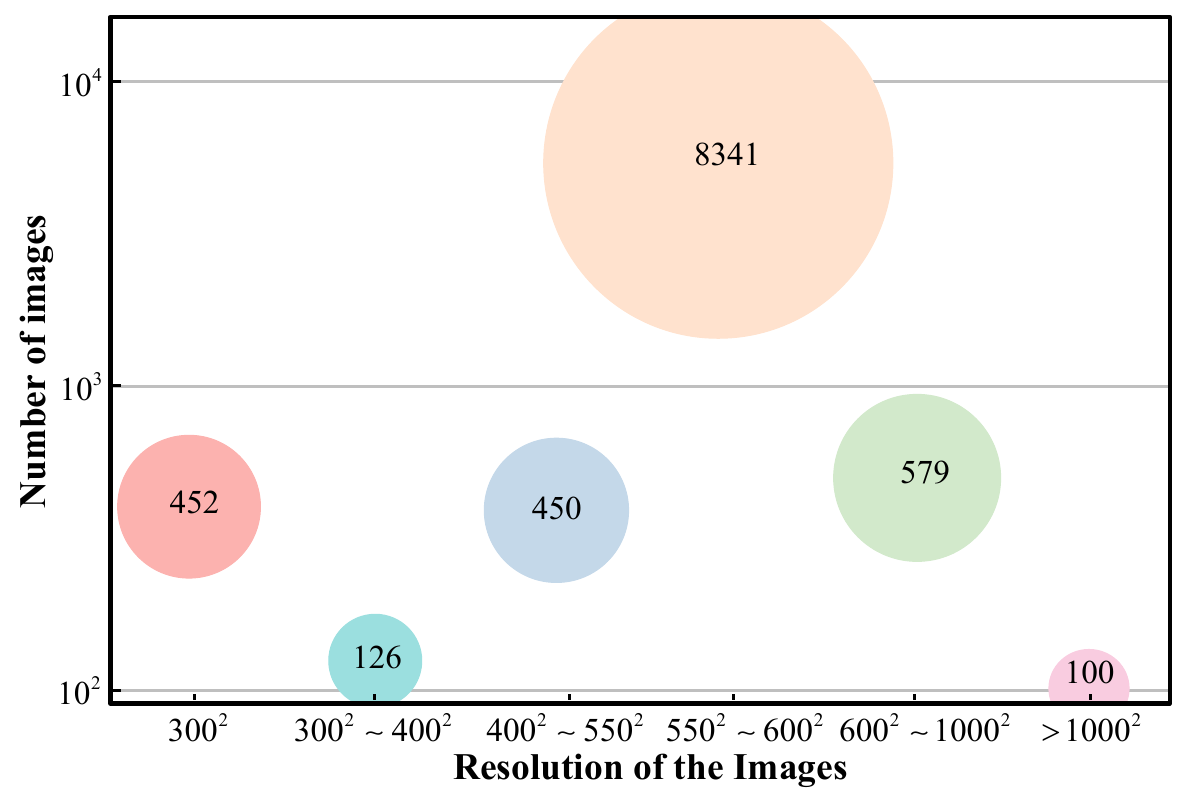}%
\vspace{-2mm}
\caption{Distribution of image resolutions in the \datasetName~dataset. 
}
\label{fig:resolution}
\vspace{-2mm}
\end{figure}

\noindent\textbf{Channel Intensity of Underwater Images.} Optical images captured in underwater environments inevitably exhibit color attenuation due to the selective absorption of different wavelengths by the water, with the red channel attenuating by an order of magnitude more than the blue and green channels \cite{colordistortion_2001_OceanSci}. As a result, underwater images typically appear bluish or greenish. To quantify this attenuation characteristic of the \datasetName, we calculated the average channel intensities and probability densities for the R, G, and B channels for each image, as shown in Fig. \ref{fig:channel}. The results show that the red channel has the lowest intensity, but also follows a similar trend to the green and blue channels.

\noindent\textbf{Size of instances.} Segmenting too small or too large instances is a common but challenging problem in the field of instance segmentation. In \cite{MSCOCO_2014_ECCV}, instances are divided into three levels: small ($area < 32^2$), medium ($32^2 \le area < 96^2$) and large ($area \ge 96^2$), where $area$~denotes the number of pixels in an instance. The number of instances of these three levels in the \datasetName~dataset are 15,800, 13,702, and 12,360, respectively, with a ratio of 1.27:1.05:1. This results in a nearly balanced distribution of instance sizes in \datasetName, which helps the network learn to segment instances of various sizes effectively. 

\noindent\textbf{Various image resolutions and image scenarios.} As shown in Fig. \ref{fig:resolution}, \datasetName~dataset includes images of various resolutions, ranging from low-resolution images captured by hand-held cameras to medium-resolution images taken by industrial equipment during underwater missions, with a small number of high-resolution images as well. This diverse resolution range is designed to meet the requirements of different tasks. 
Additionally, the \datasetName~dataset features images from shallow waters (e.g., sixth column, second row of Fig. \ref{fig:dataset_show}), images with significantly degraded quality (e.g., fourth column, second row of Fig. \ref{fig:dataset_show}), images with complex backgrounds (e.g., second column, second row of Fig. \ref{fig:dataset_show}), and images with high saturation or contrast (e.g., fourth column, first row of Fig. \ref{fig:dataset_show}). With these scenes, \datasetName~can comprehensively assess the generalizability of the model in different underwater scenarios.

\section{Method}\label{sec:method}

\begin{figure*}[!t]
\centering
\includegraphics[width=1\linewidth]{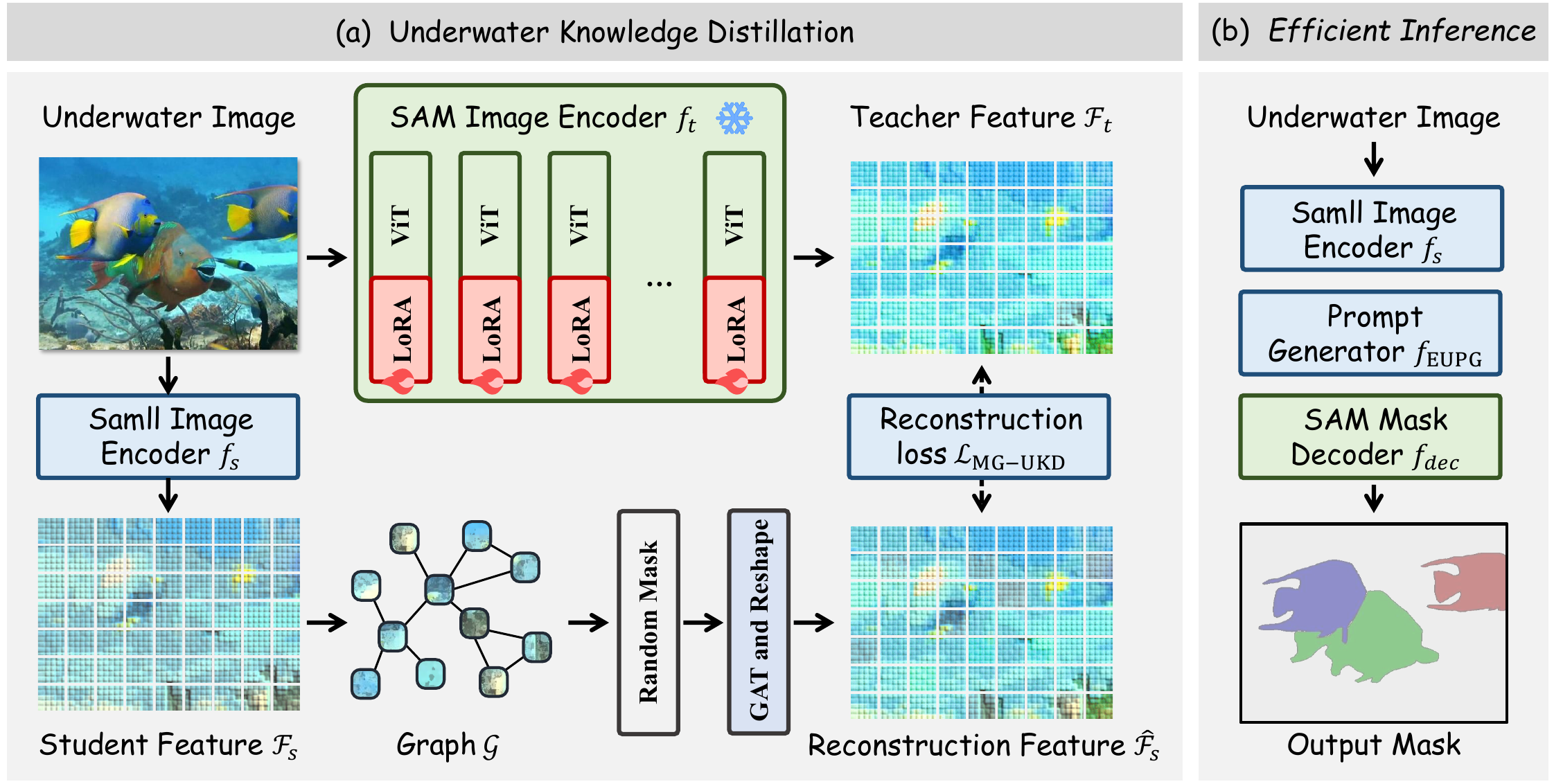}%
\caption{The overall framework of the \methodName. The left part represents the Mask GAT-based Underwater Knowledge Distillation (MG-UKD, in Section \ref{method:mgukd}), the masked features from the small image encoder are reconstructed using a GAT to align with the fine-tuned SAM encoder for efficient underwater knowledge distillation. The right part demonstrates efficient inference using the distilled model and End-to-End Underwater Prompter Generator (EUPG, in Section \ref{method:eupg}). The integration of EUPG allows the network to generate prompts directly without external input, resulting in an efficient, streamlined inference process suitable for underwater environments.}
\label{fig:framework}
\vspace{-2mm}
\end{figure*}

\subsection{Preliminary}

To begin with, we provide an overview of the SAM architecture. 
The SAM consists of three main components: an image encoder $f_{img}$, a prompt encoder $f_{prmt}$, and a mask decoder $f_{dec}$.
The image encoder $f_{img}$~leverages a Vision Transformer (ViT) trained using the Masked Autoencoder (MAE) approach \cite{MAE_2022_CVPR}.
Specifically, SAM typically employs the ViT-Huge variant.
The prompt encoder $f_{prmt}$~can handle both sparse prompts (\eg, points, boxes) $\mathcal{P}_{spare}$~and dense prompts (\eg, mask) $\mathcal{P}_{dense}$~provided by the outsiders, and converting them into prompt features. 
The mask decoder $f_{dec}$~is based on a modified Transformer decoder block \cite{Transformer_2017_NIPS}. 
Through two-way cross-attention, mask decoder $f_{dec}$~models the interaction between spare prompt features $\mathcal{F}_{spare}$~and image features $\mathcal{\tilde{F}}_{img}$, and predicts the output mask $\mathcal{M}_{out}$~with the corresponding IoU scores $\mathbf{S}_{IoU}$~through pre-inserted learnable tokens $\mathbf{T}_{token}$~(including mask token and IoU token).
The inference process of SAM can be expressed as follows:
\begin{equation}\begin{gathered}\label{eq:SAM}
\mathcal{F}_{img} = f_{img}(\mathcal{I})\text{,} \\
\mathcal{F}_{spare}, \mathcal{F}_{dense} = f_{prmt}(\mathcal{P}_{spare}, \mathcal{P}_{dense})\text{,} \\
\mathcal{\tilde{F}}_{img} = \mathcal{F}_{img} + \mathcal{F}_{dense}\text{,} \\
\mathcal{M}_{out}, \mathbf{S}_{IoU} = f_{dec}(\mathcal{\tilde{F}}_{img}, Cat(\mathbf{T}_{token}, \mathcal{F}_{spare}))\text{.}
\end{gathered}\end{equation}
where $\mathcal{I}$~denotes the input image and $Cat(\cdot)$ denotes feature combination. 
SAM2 extends SAM to video segmentation by incorporating components such as memory attention and a memory bank, but since our focus is on image segmentation, and SAM2 uses essentially the same structure as SAM in this case, with minor modifications such as using Hiera-Large \cite{Hiera_2023_PMLR}~as the image encoder, so we do not elaborate further.

\subsection{Framework of \methodName}
Figure \ref{fig:framework}~shows the overview of \methodName, which introduces two additional components into the SAM architecture: the Mask GAT-based Underwater Knowledge Distillation (MG-UKD) and the End-to-End Underwater Prompt Generator (EUPG).
MG-UKD effectively transfers knowledge from a fine-tuned SAM image encoder to a smaller, lightweight ViT-Small image encoder suitable for underwater environments.
By utilizing a Graph Attention Network (GAT)\cite{GAT_2018_ICLR}, MG-UKD aligns the output features of the smaller encoder with those from the larger SAM encoder, ensuring effective feature representation while minimizing computational requirements. 
This knowledge distillation process is reinforced by both reconstruction loss and task loss, helping the model learn accurate underwater segmentation representations.
EUPG, on the other hand, enhances efficiency by eliminating the need for external detectors and manual prompt generation.
It integrates prompt generation within the model itself, enabling an end-to-end process where underwater prompts are generated and used directly for segmentation. 
This approach improves both segmentation performance and efficiency, making it suitable for real-time, resource-constrained underwater scenarios.
We will provide a detailed explanation of these two modules in the following subsections.

\subsection{Mask GAT-based Underwater Knowledge Distillation}
\label{method:mgukd}

\begin{algorithm}[t]
\caption{Mask GAT-based Underwater Knowledge Distillation (MG-UKD)}
\label{alg:MG-UKD}
\begin{algorithmic}
\STATE {\textbf{Input:}} Transfer set $\mathcal{D}$, Teacher model $f_t$ (SAM ViT-Huge), Student model $f_s$ (ViT-Small), Layers $\mathcal{L}$~to align.
\STATE {\textbf{Output:}} Trained student model $f_s$.
\FOR{image $\mathcal{I}$ in $\mathcal{D}$}
    \STATE $\left \{ \mathcal{F}^{1}_{t}, \dots \mathcal{F}^{n}_{t} \right \} \gets f_t \left ( \mathcal{I} \right )$, $\left \{ \mathcal{F}^{1}_{s}, \dots \mathcal{F}^{n}_{s} \right \} \gets f_s \left ( \mathcal{I} \right )$;
    \FOR{layer $l$ in $\mathcal{L}$}
        \STATE Reshape $\mathcal{F}^{l}_{s}$~to Graph $\mathcal{G}^l$, and compute the first-order neighborhood $\mathcal{N}^{\,l}_i$~of each node using Equation \ref{eq:cos_similar};
        \STATE Randomly mask the features of the nodes in $\mathcal{G}^l$;
        \STATE Reconstruct the graph $\mathcal{G}^l$~as $\mathcal{G}^{l}_\mathrm{R}$~using Equation \ref{eq:GAT}-\ref{eq:update}, and reshape it as $\hat{\mathcal{F}}^{l}_s$;
        \STATE $\mathcal{L}_\mathrm{MG-UKD}^l \gets MSE(\mathcal{F}^l_t, \hat{\mathcal{F}}^l_s)$;
    \ENDFOR
    \STATE Using Loss $\mathcal{L}_\mathrm{task}+\alpha \cdot \sum_{l\in \mathcal{L}}\mathcal{L}_\mathrm{MG-UKD}^l$ to update $f_s$;
\ENDFOR
\end{algorithmic}
\end{algorithm}

Since the SAM image encoder trained on the large-scale generic image segmentation dataset SA-1B, possesses robust image feature extraction capabilities and extensive visual knowledge, it is valuable for distilling student models. However, due to the domain bias between the SA-1B dataset and the complex marine environment, the SAM image encoder lacks domain-specific expertise in the underwater environment. In addition, existing knowledge distillation approaches are not specifically designed for underwater environments and do not incorporate underwater prior knowledge in the distillation process. Therefore, directly using the SAM image encoder as a teacher model to train an underwater instance segmentation model may result in sub-optimal performance.
To address the above issues, we propose the MG-UKD.
The knowledge distillation process of MG-UKD is inspired by a characteristic of underwater images: in marine environments, objects often appear in clusters (\eg,  fish schools, coral groups), causing similar visual information to recur in different regions of the image.
Thus, compared to MGD \cite{MGD_2022_ECCV}~or SAMI \cite{EfficientSAM_2024_CVPR}, the MG-UKD approach enables the student model to reconstruct masked patches by utilising patch-like information in the underwater image through the Graph Attention Network (GAT) \cite{GAT_2018_ICLR}. This approach allows the model to better utilise the contextual similarities that exist in the underwater image during distillation.

Specifically, in this paper, we use the SAM Image Encoder fine-tuned by LoRA method \cite{LoRA_2022_ICLR}~on the UIIS10K dataset as the teacher model and the small ViT image encoder as the student model.
After that, we use each token in the output feature of the $l$-th layer $\mathcal{F}^{l}_{s} \in \mathbb{R}^{hw \times c}$ from the student model as a node of the graph $\mathcal{G}^l=\left \{ \mathbf{H}^{l}_i \in \mathbb{R}^c \middle | 1 \le i \le hw \right \}$, which does not have an excessive number of nodes since ViT will downsample the input image 16 times.
Subsequently, we connect edges based on the cosine similarity between nodes to ensure that information can propagate among similar patches. Specifically, for a given node $\mathbf{H}^{l}_i$, its first-order neighborhood $\mathcal{N}_i^{\,l}$~can be represented as:
\begin{equation}\label{eq:cos_similar}
\mathcal{N}^{\,l}_i = \left \{ \mathbf{H}^{l}_n \middle |  \dfrac{\mathbf{H}^{l}_i \cdot \mathbf{H}^{l}_n}{\max(\Vert \mathbf{H}^{l}_i \Vert _2  \cdot \Vert \mathbf{H}^{l}_n \Vert _2, \epsilon) } \ge \theta_k \right \}\text{,}
\end{equation}
where $\Vert \cdot \Vert _2$~denotes the L2 norm, $\epsilon$ is a small value to prevent division by zero, and $\theta_k$~is the similarity threshold.
In this paper, we use a dynamic threshold, setting $\theta_k$ to the cosine similarity between node $\mathbf{H}^{l}_i$ and the k-th most similar node, to prevent a node from being connected to too many or too few edges.

Then, we randomly masked most of the node features of the graph G, thus attempting to reconstruct the features $\hat{\mathcal{F}}^l_s$~through GAT and align it with the $l$-th layer features $\mathcal{F}^l_t$~of the SAM Image Encoder, thus forcing the student network to pay attention to the essential information in the image during the distillation process.
When using GAT reconstruction, we first calculate the attentional weights between the masked node $\bar{\mathbf{H}}^{l}_i$~and node $\bar{\mathbf{H}}^{l}_j$~by using the following equation:
\begin{equation}\label{eq:GAT}
a_{ij} = \frac{exp(\sigma(f_{a}(\mathbf{W}\bar{\mathbf{H}}^{l}_i\parallel \mathbf{W}\bar{\mathbf{H}}^{l}_j)))}{\sum_{n\in \mathcal{N}^{\,l}_i}\,exp(\sigma(f_{a}( \mathbf{W}\bar{\mathbf{H}}^{l}_i\parallel \mathbf{W}\bar{\mathbf{H}}^{l}_n])))}\text{,}
\end{equation}
where $\parallel$~is the concatenation operation, the shared matrix $\mathbf{W} \in \mathbb{R}^{c\times c}$ is learnable, the MLP function $f_{a} : \mathbb{R}^{2c} \to \mathbb{R}$ compute the attention coefficients for the node pairs, and $\sigma$~is the LeakyReLU function. The reconstructed node can be represented as:
\begin{equation}\label{eq:update}
\mathbf{R}^{l}_i = \sigma \left( \sum_{n\in \mathcal{N}^{\,l}_i}a_{in} \cdot \mathbf{W}\bar{\mathbf{H}}^{'}_n \right)\text{.}
\end{equation}

Finally, we reshape the reconstructed graph $\mathcal{G}^{l}_\mathrm{R}=\{ \mathbf{R}^{l}_i | 1 \le i \le hw \}$~as $\hat{\mathcal{F}}^{l}_s$~and calculate its Mean Square Error (MES) with respect to $\mathcal{F}_t$~as distillation loss. To ensure the gradual alignment of intermediate features between the student model and the teacher model, we apply MG-UKD at the 3rd, 6th, 9th, and 12th layers of them. Consequently, the final reconstruction function $\mathcal{L}_{MG-UKD}$~can be expressed as:

\begin{equation}\label{eq:MG-UKD_Loss}
\mathcal{L}_\mathrm{MG-UKD} = \sum_{l} \left( MSE(\mathcal{F}^l_t, \hat{\mathcal{F}}^l_s)\right)\text{.}
\end{equation}

\subsection{End-to-end Underwater Prompt Generator}
\label{method:eupg}

The UIIS task requires the model to recognize and segment each object in underwater images automatically.
However, existing foundational models (such as SAM \cite{SAM_2023_ICCV}, SAM2 \cite{SAM2_2024_arXiv}, or CoralSCOP \cite{CoralSCOP_2024_CVPR}) require the user to explicitly provide foreground points, bounding boxes, or masks as prompts to guide the model’s segmentation. 
A common solution is to integrate an object detection network to identify the position of the object and use the bounding box as a prompt input for the prompt encoder of these models. However, this approach not only increases the model's complexity but also limits the generated prompt embeddings to positional information alone, neglecting other essential features such as the object’s external appearance. This restriction hinders the model's optimization for underwater environments. In contrast, End-to-end Underwater Prompt Generator (EUPG) directly locates the position of instances and encodes the image features within the corresponding bounds as prompt embeddings $\mathcal{F}_{prmt}$. This enables the prompt embedding $\mathcal{F}_{prmt}$, to simultaneously incorporate both the feature and location information for each instance, enhancing model performance and adaptability in underwater tasks.

\begin{figure}[!t]
\centering
\includegraphics[width=1\linewidth]{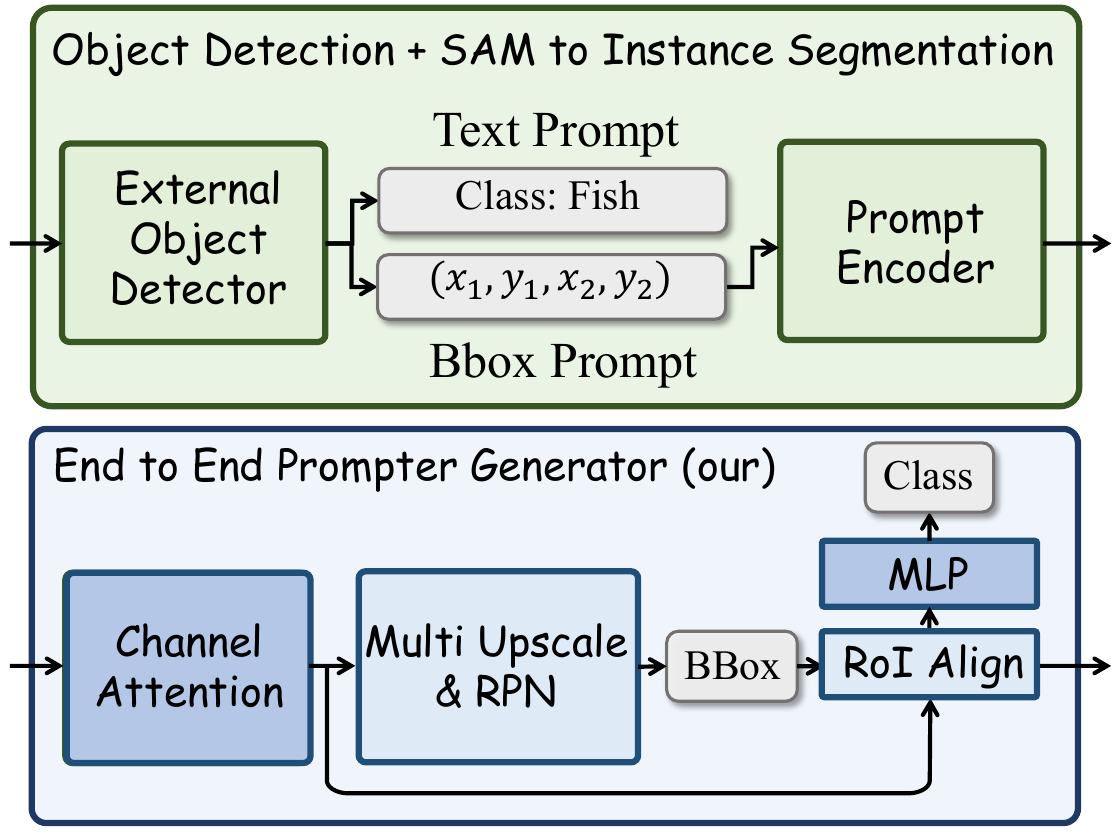}%
\vspace{-2mm}
\caption{Framework of Our End-to-end Underwater Prompt Generator.}
\vspace{-2mm}
\label{fig:pg}
\end{figure}

The structure of EUPG is shown in Figure \ref{fig:pg}.
Firstly, EUPG mitigates the light attenuation effects caused by the water medium by adjusting the weights of different channels through a simple channel attention function $f_\mathrm{CA}$, as described in the following equation:
\begin{equation}\begin{gathered}\label{eq:channel_attention}
a_{max}=Conv_{1\times1}(\sigma (Conv_{1\times1}(Pool_\mathrm{max}(F))))\text{,} \\
a_{avg}=Conv_{1\times1}(\sigma (Conv_{1\times1}(Pool_\mathrm{avg}(F))))\text{,} \\
\mathcal{\tilde{F}}_{img} = f_\mathrm{CA}(\mathcal{F}_{img})=Sigmoid(a_{max}+a_{avg}) \cdot \mathcal{F}_{img} \text{,}
\end{gathered}\end{equation}
where $Pool_\mathrm{max}$~and $Pool_\mathrm{avg}$~denote global maximum pooling and global average pooling respectively.
In fact, this channel attention function can be regarded as an adapter \cite{USISSAM_2024_ICML}~attached to UWSAM after the image encoder to enhance the prompt embedding representation generated afterward by dynamically adjusting the channel information.

Furthermore, to assist the model in detecting and extracting prompt embeddings for instances of varying sizes, we apply 2x and 4x upsampling to the features after channel attention adjustment. 
The resulting feature maps are then fed into a Region Proposal Network (RPN) \cite{FastRCNN_2015_ICCV}~for efficient object localization.
Subsequently, we map the image features corresponding to the position of each instance to $14\times 14\times c$~through RoIAlign intervals and will be mapped to the final features through convolution, flatten, and MLP  operations. Specifically, the above process can be represented as:
\begin{equation}\begin{gathered}\label{eq:RoIAlign}
\mathcal{F}_{roi} = RoIAlign(\mathcal{F}_{img} + \mathcal{F}_{pe}, \mathcal{B})\text{,} \\
\mathcal{F}_{prmt} = MLP(Flatten(Conv_{3\times3}(\mathcal{F}_{roi})))\text{,} \\
\end{gathered}\end{equation}
where $\mathcal{F}_{pe}$ denotes the positional encoding and $\mathcal{B}$~denotes the location result of the RPN. With the introduction of EUPG, the inference process of UWSAM can be expressed as follows:
\begin{equation}\begin{gathered}\label{eq:EUPG_Pipline}
\mathcal{F}_{img} = f_{s}(\mathcal{I})\text{,} \\
\mathcal{F}_{prmt}, \mathbf{C}_{cls} = f_\mathrm{EUPG}(\mathcal{F}_{img})\text{,} \\
\mathcal{M}_{out}, \mathbf{S}_{IoU} = f_{dec}(\mathcal{F}_{img}, Cat(\mathbf{T}_{token}, \mathcal{F}_{prmt}))\text{.}
\end{gathered}\end{equation}
where $\mathbf{C}_{cls}$~is the classification result for the corresponding instance
Thus, UWSAM can achieve end-to-end efficient underwater instance segmentation through the EUPG module without using SAM's Prompt encoder

\subsection{Loss Function}
We define our training loss function as follows:
\begin{equation}\label{eq:loss}
\mathcal{L}=\mathcal{L}_\mathrm{Task}+\alpha \cdot \mathcal{L}_\mathrm{MG-UKD},
\end{equation}
where $\mathcal{L}_\mathrm{Task}$~is the loss corresponding to the to the instance segmentation and $\mathcal{L}_\mathrm{MG-UKD}$~is the reconstruction loss generated during distillation using the MG-UKD algorithm.
The hyperparameter $\alpha$~controls the balance between these two components and is set to $2 \times 10^{-5}$ in this work.

The task loss $\mathcal{L}_\mathrm{Task}$~of the \methodName~is similar to the Mask RCNN \cite{MaskRCNN_2017_ICCV}~and consists of several components \ie, localization loss, classification loss, and segmentation loss. Therefore, the task loss $\mathcal{L}_\mathrm{Task}$~can be expressed as:
\begin{equation}\label{eq:task_loss}
\mathcal{L}_\mathrm{Task}=\mathcal{L}_\mathrm{cls}+\mathcal{L}_\mathrm{rpn}+\mathcal{L}_\mathrm{seg},
\end{equation}
where $\mathcal{L}_\mathrm{cls}$~and $\mathcal{L}_\mathrm{rpn}$~is the sum of Cross Entropy (CE) loss about classification and Smooth L1 loss about localization in the EUPG, and $\mathcal{L}_\mathrm{seg}$~are the CE loss in the SAM decoder.

\begin{table*}[!t]
    \caption{Quantitative comparisons on the \datasetName~ datasets. ViT-Huge* indicates that the backbone is frozen during training, while LoRA \cite{LoRA_2022_ICML}~and Adapter \cite{Adapter_ICML}~refer to efficient parameter fine-tuning methods applied to the corresponding backbone during training.}
    \centering
    \renewcommand{\arraystretch}{1.4}
    \begin{tabular}{c|c|c|ccc|ccc|c}
    \hline
    \hline
    Model & Backbone & Epoch & $\mathrm{mAP^{b}}$ & $\mathrm{AP^{b}_{50}}$ & $\mathrm{AP^{b}_{75}}$ & $\mathrm{mAP^{s}}$ & $\mathrm{AP^{s}_{50}}$ & $\mathrm{AP^{s}_{75}}$ & Params\\
    \hline
    Mask RCNN\cite{MaskRCNN_2017_ICCV} & ResNet-101 & 24 & 38.4 & 54.4 & 42.3 & 35.8 & 53.6 & 40.2&60.09 M \\
    Cascade Mask RCNN\cite{Cai_2019_crcnn} & ResNet-101 & 24 & 40.8 & 55.1 & 44.1 & 36.6 & 53.9 & 40.5& 91.60 M\\
    Mask Scoring RCNN\cite{huang2019msrcnn} & ResNet-101 & 24 & 38.8 & 54.5 & 42.7 & 37.3 & 53.7 & 41.8 &75.68 M\\
    YOLACT\cite{bolya2019yolactreal} & ResNet-101 & 55 & 33.7 & 52.3 & 37.9 & 35.0 & 50.9 & 37.9 & 51.29 M\\
    PointRend\cite{kirillov2020pointrend} & ResNet-50 & 36 & 37.0 & 54.0 & 40.6 & 37.3 & 53.0 & 41.3 & 53.66 M\\
    QueryInst\cite{QueryInst_2021_ICCV} & ResNet-101 & 24 & 36.8 & 48.8 & 39.8 & 33.9 & 48.8 & 38.7 & 182.60 M\\
    Mask2Former\cite{Mask2Former_2022_CVPR} & Swin-Base & 24 & 26.6 & 34.6 & 26.0 & 31.7 & 45.2 & 33.2 & 101.93 M\\
    CondInst\cite{tian2020conditional} & ResNet-101 & 24 & 33.1 & 46.1 & 34.8 & 32.5 & 45.4 & 34.9 & 50.54 M\\
    RTMDet\cite{lyu2022rtmdet} & CSPNeXt & 24 & 20.1 & 29.0 & 21.4 & 19.0 & 27.4 & 20.5 & 97.88 M\\
    BoxInst\cite{tian2020box} & ResNet-101 & 24 & 33.2 & 47.4 & 35.7 & 26.4 & 41.2 & 29.1 & 51.47 M\\
    ConvNeXt-V2\cite{woo2023convnext} & ConvNeXt-V2-B & 24 & 35.8 & 49.7 & 40.6 & 35.2 & 49.3 & 39.3 & 102.72 M\\
    WaterMask\cite{WaterMask_2023_ICCV} & ResNet-101 & 24 & 35.1 & 52.0 & 39.0 & 37.4 & 51.6 & 41.7 & 63.80 M\\
    MaskRCNN\cite{MaskRCNN_2017_ICCV}+EfficientSAM\cite{EfficientSAM_2024_CVPR}& ViT-Small & 24 & 40.2 & 55.1 & 43.9 & 36.6 & 52.1 & 40.2 & \textbf{42.21 M}\\
    \rowcolor[RGB]{240,250,239}\methodName-Student & ViT-Small & 24 & 41.4 & 56.3 & 45.7 & 38.7 & 54.3 & 42.1 & \underline{47.39 M}\\
    \hline
    FasterRCNN\cite{FastRCNN_2015_ICCV}+SAM\cite{SAM_2023_ICCV}& ResNet-101+ViT-Huge* & 24 & 31.0 & 46.1 & 34.7 & 35.4 & 45.7 & 37.9 & 669.07 M\\
    MaskRCNN\cite{MaskRCNN_2017_ICCV}+SAM\cite{SAM_2023_ICCV}& ViT-Huge (LoRA) & 24 & \underline{44.7}  & \underline{58.3} & \underline{49.2} & \underline{42.5} & \underline{57.9} & \underline{48.4}& 627.13 M\\
    MaskRCNN\cite{MaskRCNN_2017_ICCV}+SAM2\cite{SAM2_2024_arXiv} & Hiera-Large (LoRA) & 24 & 36.9 & 50.9 & 40.5 & 40.6 & 50.8 & 44.0 & 224.24 M\\
    USIS-SAM\cite{USISSAM_2024_ICML} & ViT-Huge (Adapter) & 24 & 35.8 & 53.6 & 40.6 & 39.8 & 52.0 & 42.6 & 665.80 M\\
    \rowcolor[RGB]{240,250,239}\methodName-Teacher & ViT-Huge (LoRA)  & 24 & \textbf{45.8} & \textbf{60.2} & \textbf{50.6} & \textbf{44.6} & \textbf{58.9} & \textbf{49.2} & 632.30 M\\
    \hline
    \hline
    \end{tabular}
    \label{tab:uiis10k}
\end{table*}

\section{EXPERIMENTS}
We comprehensively evaluate state-of-the-art instance segmentation methods with our proposed \methodName~on the UIIS10K, UIIS \cite{WaterMask_2023_ICCV}, and USIS10K \cite{USISSAM_2024_ICML}~datasets to analyze their performance in underwater object detection and instance segmentation tasks. For evaluation metrics, we refer to MaskRCNN \cite{MaskRCNN_2017_ICCV}~and use standard bbox AP metrics ($\mathrm{mAP^{b}}\text{, }\mathrm{AP^{b}_{50}}\text{, }\mathrm{AP^{b}_{75}}$) and mask AP metrics ($\mathrm{mAP^{s}}\text{, }\mathrm{AP^{s}_{50}}\text{, }\mathrm{AP^{s}_{75}}$) to evaluate the model's object detection and instance segmentation capabilities. 
Furthermore, we conduct comparative experiments against with other knowledge distillation methods on the UIIS10K dataset. Following MGD \cite{MGD_2022_ECCV} and EfficientSAM\cite{EfficientSAM_2024_CVPR}, we use mAP for Small, Medium, and Large objects ($\mathrm{AP^{s}_{S}}\text{, }\mathrm{AP^{s}_{M}}\text{, }\mathrm{AP^{s}_{L}}$) as the evaluation metric.
Finally, we also performed ablation studies on other modules to demonstrate the effectiveness of the proposed approach.

\subsection{Implementation Details}
We implement \methodName~and other comparison algorithms using PyTorch and MMDetection \cite{mmdetection} framework. 
All backbones and hyperparameters of the methods and comparison algorithms are the same as original paper, except for our newly designed parts.
In the MK-UKD, we employ ViT-Small as the student network backbone and a LoRA-tuned SAM image encoder (ViT-Huge) as the teacher model backbone. 
Following MGD \cite{MGD_2022_ECCV}, we set the feature masking ratio to 0.65 during distillation and employ a two-layer Graph Attention Network (GAT) \cite{GAT_2018_ICLR}~with four attention heads to reconstruct the masked features. For graph construction, we follow WaterMask \cite{WaterMask_2023_ICCV}~and set $k=11$ in Equation \ref{eq:cos_similar}.

For model optimization, We trained our method on 2 NVIDIA 4090 GPUs using the AdamW optimizer for 24 epochs with a starting learning rate of 2e-4 and weight decay of 5e-2. For data expansion, we used random flipping, random scaling, and random cropping in all methods to ensure fair comparisons. All comparison algorithms were retrained on the underwater dataset to adapt to the underwater environment. 

\subsection{Datasets}
We conduct experiments on three datasets: our UIIS10K dataset, the UIIS dataset \cite{WaterMask_2023_ICCV}, and the USIS10K dataset \cite{USISSAM_2024_ICML}. The first two are used for underwater instance segmentation, while the latter is used for underwater salient instance segmentation. 
We will describe each dataset in detail below.

\noindent\textbf{UIIS10K Dataset}.
We evaluate our method and state-of-the-art instance segmentation methods in underwater environments using the UIIS10K dataset. We follow the settings in Section \ref{subsec:dataset}~and use 8,083 images for training and 2,010 images for validation and testing.

\noindent\textbf{UIIS Dataset}.
UIIS dataset \cite{WaterMask_2023_ICCV}~is the first large-scale general underwater image instance segmentation dataset. It comprises approximately 25,000 images collected from various domains, which were filtered using the Underwater Color Image Quality Evaluation (UCIQE) \cite{UCIQE_2015_TIP} and Underwater Image Quality Measurement (UIQM) \cite{UIQM_2016_JOE} metrics to select 4,628 high-quality underwater images. These images are meticulously annotated into seven categories, with 3,937 images used for training and 691 images used for validation and testing.

\noindent\textbf{USIS10K Dataset}.
The USIS10K dataset \cite{USISSAM_2024_ICML}~is the first large-scale underwater salient instance segmentation dataset, consisting of 10,632 pixel-level annotated images for seven underwater categories. The dataset is split into training, validation, and test sets following a 7:1.5:1.5 ratio, resulting in 7,442 images for training, 1,594 for validation, and 1,597 for testing. In this work, we evaluate our method and state-of-the-art instance segmentation approaches on USIS10K for the underwater salient instance segmentation task to assess the generalizability of each method.

\subsection{Main Results}
\noindent\textbf{Comparison With Instance Segmentation Model}.
In this subsection, we evaluate \methodName~against state-of-the-art instance segmentation approaches, including one-stage methods \cite{bolya2019yolactreal, tian2020conditional, lyu2022rtmdet, tian2020box}, two-stage methods \cite{MaskRCNN_2017_ICCV, Cai_2019_crcnn, huang2019msrcnn, kirillov2020pointrend, woo2023convnext, WaterMask_2023_ICCV}, and query-based methods \cite{QueryInst_2021_ICCV, Mask2Former_2022_CVPR}.
Firstly, Table \ref{tab:uiis10k}~presents the performance comparison between our method and other models on the UIIS10K dataset. In comparison to the general baseline for instance segmentation, Mask R-CNN \cite{MaskRCNN_2017_ICCV},  the distilled UWSAM-Student model achieves improvements of 3.0 AP and 2.9 AP in $\mathrm{mAP^b}$~and $\mathrm{mAP^s}$, respectively. This improvement suggests that MU-UKD enhances UWSAM-Student’s feature extraction capability for underwater images through knowledge distillation.
Furthermore, when compared to the well-established query-based instance segmentation method YOLCAT \cite{bolya2019yolactreal}, UWSAM-Student outperforms it by 3.7 AP, 3.4 AP, and 4.3 AP in $\mathrm{mAP^s}$, $\mathrm{AP^s_{50}}$, and $\mathrm{AP^s_{75}}$, respectively. This benefit comes from the powerful instance segmentation capability brought by the SAM framework.

\begin{table*}[t]
    \caption{Quantitative comparisons on the UIIS datasets. ViT-Huge* indicates that the backbone is frozen during training, while LoRA \cite{LoRA_2022_ICML}~and Adapter \cite{Adapter_ICML}~refer to efficient parameter fine-tuning methods applied to the corresponding backbone during training.}
    \centering
    \renewcommand{\arraystretch}{1.4}
    \begin{tabular}{c|c|c|ccc|ccc|c}
    \hline\hline
    Model & Backbone & Epoch & $\mathrm{mAP^{b}}$ & $\mathrm{AP^{b}_{50}}$ & $\mathrm{AP^{b}_{75}}$ & $\mathrm{mAP^{s}}$ & $\mathrm{AP^{s}_{50}}$ & $\mathrm{AP^{s}_{75}}$ & Params \\
    \hline
    Mask RCNN\cite{MaskRCNN_2017_ICCV} & ResNet-101 & 24 & 27.2 & 45.7 & 28.5 & 24.4 & 45.1 & 23.4 & 60.08 M\\
    Cascade Mask RCNN\cite{Cai_2019_crcnn} & ResNet-101 & 24 & 28.0 & 44.7 & 29.6 & 24.7 & 43.3 & 25.8 & 91.59 M \\
    Mask Scoring RCNN\cite{huang2019msrcnn} & ResNet-101 & 24 & 26.2 & 42.7 & 28.3 & 24.2 & 42.3 & 24.5 & 75.59 M\\
    YOLACT\cite{bolya2019yolactreal} & ResNet-101 & 55 & 21.2 & 39.3 & 20.9 & 18.5 & 36.2 & 17.8 & 51.27 M\\
    QueryInst\cite{QueryInst_2021_ICCV} & ResNet-101 & 24 & 22.1 & 36.2 & 22.4 & 20.5 & 34.6 & 21.3 & 182.59 M\\
    Mask2Former\cite{Mask2Former_2022_CVPR} & Swin-Base & 24 & 11.1 & 19.8 & 10.6 & 17.6 & 31.5 & 17.7 &  101.93 M \\
    CondInst\cite{tian2020conditional} & ResNet-101 & 24 & 26.3 & 43.0 & 28.2 & 25.7 & 42.8 & 25.7 & 50.53 M\\
    BoxInst\cite{tian2020box} & ResNet-101 & 24 & 25.1 & 42.2 & 25.3 & 19.2 & 38.7 & 16.3 & 51.47 M\\
    ConvNeXt-V2\cite{woo2023convnext} & ConvNeXt-V2-B & 24 & 24.5 & 39.5 & 27.1 & 22.7 & 38.6 & 23.8 & 102.71 M\\
    WaterMask\cite{WaterMask_2023_ICCV} & ResNet-101 & 24 & 25.9 & 43.0 & 26.8 & 25.8 & 42.8 & 27.3 & 63.78 M \\
    MaskRCNN\cite{MaskRCNN_2017_ICCV}+EfficientSAM\cite{EfficientSAM_2024_CVPR}& ViT-Small & 24 & 28.2 & 43.6 & 31.1 & 25.2 & 42.1 & 24.5 & \textbf{42.19 M}\\
    \rowcolor[RGB]{240,250,239}\methodName-Student & ViT-Small & 24 & 31.2 & \underline{49.4} & 33.4 & 26.3 & 44.6 & 26.2 & \underline{47.39 M}\\
    \hline
    FasterRCNN\cite{FastRCNN_2015_ICCV}+SAM\cite{SAM_2023_ICCV}& ResNet-101+ViT-Huge* & 24 & 23.5 & 38.4 & 24.8 & 22.4 & 36.6 & 23.9 & 669.07 M\\
    MaskRCNN\cite{MaskRCNN_2017_ICCV}+SAM\cite{SAM_2023_ICCV}& ViT-Huge (LoRA) & 24 & \underline{32.5} & 48.0 & \underline{37.3} & 29.9 & \underline{47.8} & 33.1 & 632.23 M\\
    MaskRCNN\cite{MaskRCNN_2017_ICCV}+SAM2\cite{SAM2_2024_arXiv} & Hiera-Large (LoRA) & 24 & 27.7 & 44.1 & 29.1 & \underline{30.2} & 43.6 & \underline{33.5} & 224.23 M\\
    USIS-SAM\cite{USISSAM_2024_ICML} & ViT-Huge (Adapter) & 24 & 28.7 & 48.5 & 31.8 & 30.1 & 46.7 & 32.9 & 665.78 M\\
    \rowcolor[RGB]{240,250,239}\methodName-Teacher & ViT-Huge (LoRA)  & 24 & \textbf{34.3} & \textbf{50.6} & \textbf{37.8} & \textbf{31.9} & \textbf{48.2} & \textbf{34.9} & 632.30 M\\
    \hline\hline
    \end{tabular}
    \label{tab:uiis}
\end{table*}

Moreover, in contrast to WaterMask, a two-stage method specifically designed for underwater environments, UWSAM-Student surpasses it by 6.3 AP, 4.3 AP, and 6.7 AP in $ \mathrm{mAP^b} $, $ \mathrm{AP^b_{50}} $, and $ \mathrm{AP^b_{75}} $, respectively, as well as by 1.3 AP, 2.7 AP, and 0.4 AP in $ \mathrm{mAP^s} $, $ \mathrm{AP^s_{50}} $, and $ \mathrm{AP^s_{75}} $, respectively. These results indicate that UWSAM-Student achieves superior underwater instance object detection performance compared to WaterMask, primarily due to the strong localization capability introduced by the EUPG module.
Finally, it is important to note that UWSAM-Student has the second smallest number of parameters among all compared methods, with only 47.39 M. This makes it more suitable for deployment on memory-constrained edge computing devices, such as autonomous underwater vehicles and remotely operated underwater robots.

We also present the performance of UWSAM on the UIIS dataset and the USIS10K dataset in Tables \ref{tab:uiis}~and \ref{tab:usis10k}, respectively. As shown in Table \ref{tab:uiis}, UWSAM-Student outperforms Mask R-CNN and WaterMask by 4.0 AP and 5.3 AP in $ \mathrm{mAP^b} $, and by 1.9 AP and 0.5 AP in $ \mathrm{mAP^s} $, respectively.
Similarly, in Table \ref{tab:usis10k}, UWSAM-Student demonstrates outstanding performance in the underwater salient instance segmentation task. Specifically, it surpasses Mask R-CNN and WaterMask by 2.9 AP and 5.6 AP in $ \mathrm{mAP^b} $, and by 1.4 AP and 1.1 AP in $ \mathrm{mAP^s} $, respectively.
This demonstrates the generalization capability of UWSAM-Student across different datasets and tasks.

\begin{table*}[t]
    \caption{Quantitative comparisons on the USIS10K datasets. ViT-Huge* indicates that the backbone is frozen during training, while LoRA \cite{LoRA_2022_ICML}~and Adapter \cite{Adapter_ICML}~refer to efficient parameter fine-tuning methods applied to the corresponding backbone during training.}
    \centering
    \renewcommand{\arraystretch}{1.4}
    \begin{tabular}{c|c|c|ccc|ccc|c}
    \hline\hline
    Model & Backbone & Epoch & $\mathrm{mAP^{b}}$ & $\mathrm{AP^{b}_{50}}$ & $\mathrm{AP^{b}_{75}}$ & $\mathrm{mAP^{s}}$ & $\mathrm{AP^{s}_{50}}$ & $\mathrm{AP^{s}_{75}}$  & Params\\
    \hline
    Mask RCNN\cite{MaskRCNN_2017_ICCV} & ResNet-101 & 24 & 41.9 & 59.1 & 47.2 & 39.6 & 58.7 & 45.0 & 60.08 M\\
    Cascade Mask RCNN\cite{Cai_2019_crcnn} & ResNet-101 & 24 & 43.1 & 57.9 & 47.2 & 38.9 & 57.0 & 43.6 & 91.59 M\\
    Mask Scoring RCNN\cite{huang2019msrcnn} & ResNet-101 & 24 & 43.0 & 58.7 & 48.7 & 40.7 & 57.9 & 46.2 & 75.59 M\\
    YOLACT\cite{bolya2019yolactreal} & ResNet-101 & 55 & 37.8 & 56.4 & 42.8 & 37.6 & 55.5 & 41.7 & 51.27 M \\
    QueryInst\cite{QueryInst_2021_ICCV} & ResNet-101 & 24 & 40.9 & 55.0 & 45.6 & 37.9 & 54.7 & 43.3 & 182.59 M\\
    Mask2Former\cite{Mask2Former_2022_CVPR} & Swin-Base & 24 & 23.6 & 33.6 & 24.3 & 33.4 & 50.0 & 35.5 & 101.93 M\\
    CondInst\cite{tian2020conditional} & ResNet-101 & 24 & 40.8 & 56.6 & 45.9 & 38.9 & 55.8 & 43.1 & 50.53 M\\
    RTMDet\cite{lyu2022rtmdet} & CSPNeXt & 24 & 24.6 & 34.2 & 26.8 & 23.0 & 33.5 & 26.3 & 97.88 M\\
    BoxInst\cite{tian2020box} & ResNet-101 & 24 & 38.5 & 54.4 & 42.8 & 30.7 & 52.1 & 32.5 & 51.47 M\\
    ConvNeXt-V2\cite{woo2023convnext} & ConvNeXt-V2-B & 24 & 40.9 & 56.9 & 45.7 & 39.5 & 55.9 & 44.5 & 102.71 M\\
    WaterMask\cite{WaterMask_2023_ICCV} & ResNet-101 & 24 & 39.2 & 56.9 & 45.4 & 39.9 & 56.4 & 45.8 & 63.78 M \\
    MaskRCNN\cite{MaskRCNN_2017_ICCV}+EfficientSAM\cite{EfficientSAM_2024_CVPR}& ViT-Small & 24 & 41.8 & 54.6 & 47.3 & 39.5 & 54.9 & 45.9 & \textbf{42.19 M}\\
    \rowcolor[RGB]{240,250,239}\methodName-Student & ViT-Small & 24 & 44.8 & 59.8 & \underline{50.8} & 41.0 & 57.6 & 47.2 & \underline{47.39 M}\\
    \hline
    FasterRCNN\cite{FastRCNN_2015_ICCV}+SAM\cite{SAM_2023_ICCV}& ResNet-101+ViT-Huge* & 24 & 39.0 & 54.4 & 43.1 & 37.0 & 52.9 & 42.1 & 669.07 M\\
    MaskRCNN\cite{MaskRCNN_2017_ICCV}+SAM\cite{SAM_2023_ICCV}& ViT-Huge (LoRA) & 24 & \underline{45.2}  & 59.2 & 50.7 & 43.3 & 58.8 &49.4& 627.13 M\\
    MaskRCNN\cite{MaskRCNN_2017_ICCV}+SAM2\cite{SAM2_2024_arXiv} & Hiera-Large (LoRA) & 24 & 42.6 & 57.7 & 46.7 & \underline{44.7} & 57.5 & 51.1 & 224.23 M \\
    USIS-SAM\cite{USISSAM_2024_ICML} & ViT-Huge (Adapter) & 24 & 42.6 & \underline{61.4} & 49.0 & 43.5 & \underline{60.7} & \underline{49.6} & 665.78 M\\
    \rowcolor[RGB]{240,250,239}\methodName-Teacher & ViT-Huge (LoRA)  & 24 & \textbf{48.9} & \textbf{64.1} & \textbf{55.1} & \textbf{46.0} & \textbf{61.7} & \textbf{51.7} & 632.30 M\\
    \hline\hline
    \end{tabular}
    \label{tab:usis10k}
\end{table*}

\noindent\textbf{Comparison With Segment Anything Model}. 
In this subsection, we compare UWSAM with SAM \cite{SAM_2023_ICCV}, SAM2 \cite{SAM2_2024_arXiv}, and EfficientSAM \cite{EfficientSAM_2024_CVPR}.
Since SAM and its variants require user-provided point or bounding box prompts for inference, we follow the approaches in RSPrompter \cite{rsprompter_2024_chen}~and USIS-SAM \cite{USISSAM_2024_ICML}~to design two SAM-based pipelines for comparison with UWSAM. Specifically, FasterR-CNN+SAM introduces an additional Faster R-CNN for object detection, whose outputs are then fed into SAM for instance segmentation. Meanwhile, MaskR-CNN+SAM replaces the backbone of Mask R-CNN with the SAM Image Encoder, enabling the object detection and instance segmentation components to share a common backbone for joint optimization. Similarly, MaskR-CNN+SAM2 and MaskR-CNN+EfficientSAM follow the same principle as MaskR-CNN+SAM. In addition, since USIS-SAM \cite{USISSAM_2024_ICML}~enables end-to-end instance segmentation, we did not make additional modifications to it.

\begin{figure}[!t]
\centering
\includegraphics[width=1\linewidth]{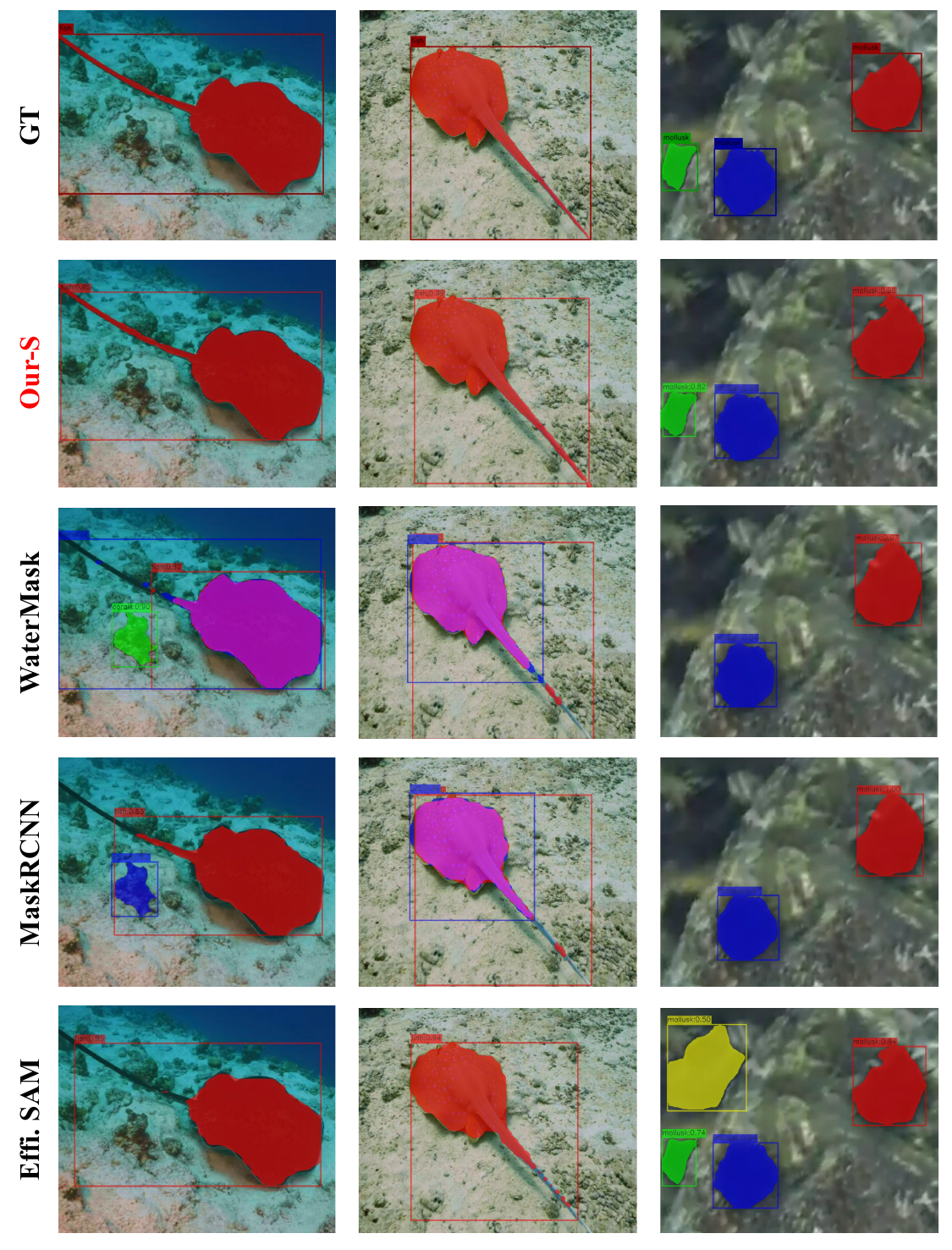}%
\vspace{-2.5mm}
\caption{Comparison of \methodName-Student (Our-S) with other instance segmentation methods \cite{MaskRCNN_2017_ICCV, WaterMask_2023_ICCV, EfficientSAM_2024_CVPR}~on the UIIS10K dataset. Effi. SAM denotes MaskRCNN\cite{MaskRCNN_2017_ICCV} + SAM\cite{EfficientSAM_2024_CVPR}. 
}
\vspace{-2.5mm}
\label{fig:stud_show}
\end{figure}

As shown in Table \ref{tab:uiis10k}, compared with FasterR-CNN+SAM, UWSAM-Student improves by 10.4 and 3.3 percentage points in $ \mathrm{mAP^b} $ and $ \mathrm{mAP^s} $ metrics, respectively. This is mainly due to the fact that UWSAM improves the performance of the model by aggregating the target detection directly to the model in, and jointly optimising the target detection with instance segmentation.
In comparison with USIS-SAM \cite{USISSAM_2024_ICML}, a SAM variant specifically designed for underwater environments, UWSAM-Student achieves comparable performance in $ \mathrm{mAP^s} $ while using only approximately 7\% of its parameters. Furthermore, it surpasses USIS-SAM by 5.6 AP and 2.2 AP in $ \mathrm{mAP^b} $ on the UIIS10K and USIS10K datasets \cite{USISSAM_2024_ICML}, respectively. This performance advantage is attributed to MG-UKD, which enhances UWSAM-Student through knowledge distillation, and EUPG, which provides the network with strong localization capabilities.
In comparison with MaskR-CNN+SAM, the UWSAM-Teacher, which has a similar number of parameters, surpasses it by 1.7 AP in $ \mathrm{mAP^b} $ and 2.0 AP in $ \mathrm{mAP^s} $. This improvement is primarily attributed to our modification of SAM’s prompt encoder into EUPG within the pipeline, enabling it to simultaneously perform localization and prompt encoding. As a result, the model can be optimized in an end-to-end manner during training.
And when compared with the baseline MaskRCNN, UWSAM-Teacher leads by 7.4, 7.1, and 7.0 APs in $ \mathrm{mAP^b} $~for the three datasets and by 8.8, 7.5, and 6.4 APs in $ \mathrm{mAP^s} $. This implies that in some offline underwater analysis tasks that do not require real-time, UWSAM-Teacher has great potential.

\begin{figure}[!t]
\centering
\includegraphics[width=1\linewidth]{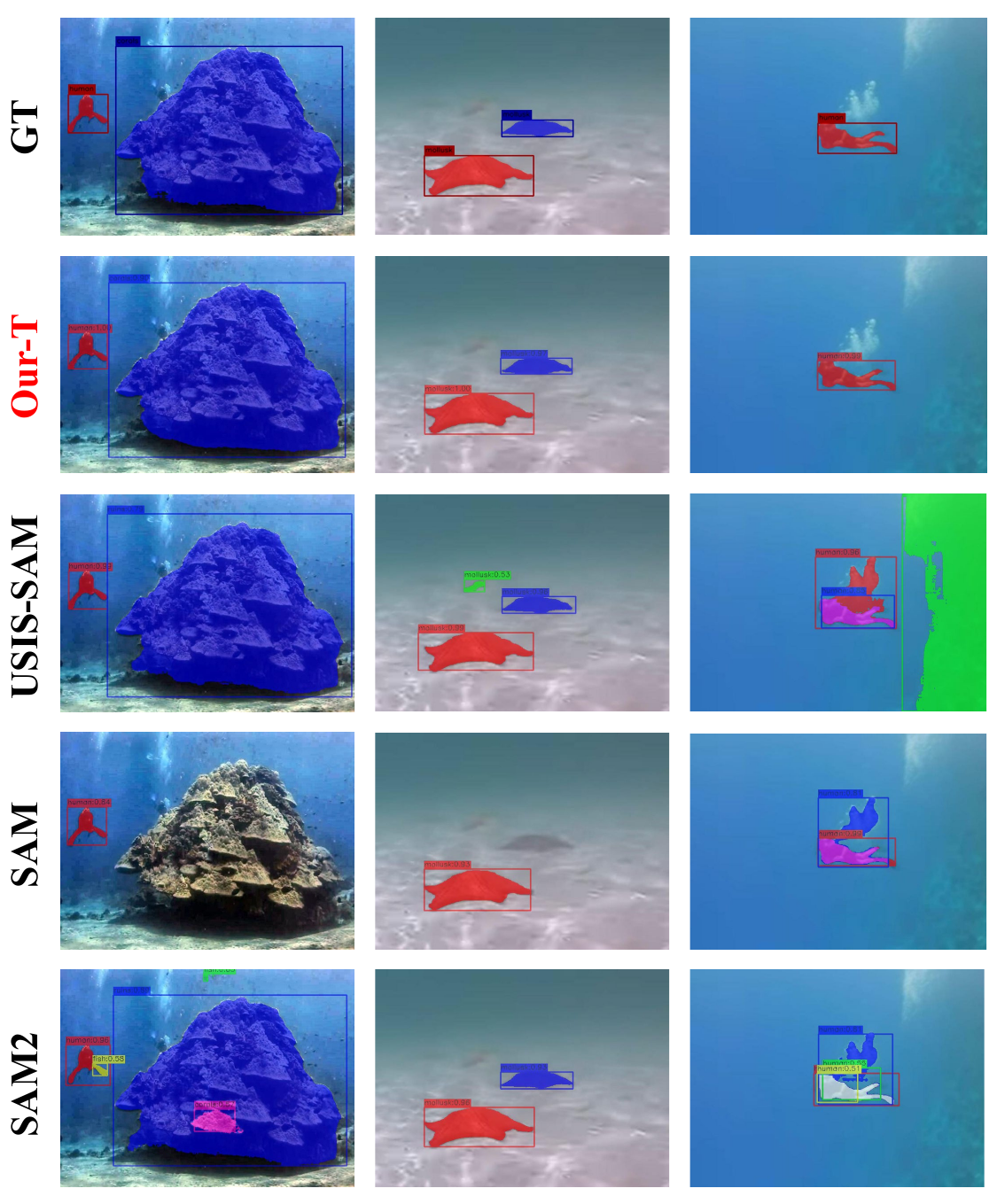}%
\vspace{-2.5mm}
\caption{Comparison of \methodName-Teacher (Our-T) with other Segment Anything methods \cite{SAM_2023_ICCV, SAM2_2024_arXiv, USISSAM_2024_ICML}~on the UIIS10K dataset. SAM / SAM2 denotes  MaskRCNN\cite{MaskRCNN_2017_ICCV} + SAM \cite{SAM_2023_ICCV} / SAM2 \cite{SAM2_2024_arXiv}.  
}
\label{fig:teach_show}
\vspace{-2.5mm}
\end{figure}


\subsection{Ablation Study}
To systematically evaluate the contributions of core components in \methodName, we conduct ablation experiments on the \datasetName~dataset.

\noindent\textbf{Mask GAT-based Underwater Knowledge Distillation}.
We first compare MG-UKD with several widely used knowledge distillation methods, including OFD \cite{OFD_2019_ICCV}, MGD \cite{MGD_2022_ECCV}, and SAMI (used in EfficientSAM) \cite{EfficientSAM_2024_CVPR}, to validate the effectiveness of our proposed approach.
Additionally, following MGD \cite{MGD_2022_ECCV} and SAMI \cite{EfficientSAM_2024_CVPR}, we also use mAP for Small, Medium, and Large objects ($\mathrm{AP^{s}{S}}$, $\mathrm{AP^{s}{M}}$, $\mathrm{AP^{s}_{L}}$) as the evaluation metric.
For a fair comparison, we use the UWSAM-Teacher, which is trained on the UIIS10K dataset, as the teacher model for all distillation methods. Each student model use ViT-Small as the backbone. All models are trained for 24 epochs using identical hyperparameters across methods to ensure consistency.

As shown in Table \ref{tab:as_mgukd}, compared to the well-established feature-based distillation method OFD \cite{OFD_2019_ICCV}, MG-UKD achieves improvements of 0.5, 0.1, and 0.8 AP in mAP, AP$_{50}$, and AP$_{75}$, respectively. Notably, the larger gain under the stricter AP$_{75}$ metric suggests that MG-UKD enhances the student model’s ability to generate more accurate and detailed predictions. This improvement can be attributed to MG-UKD’s strategy of guiding the student to recover masked features, thereby improving its representational capacity.
Furthermore, in comparison to MGD \cite{MGD_2022_ECCV}~and SAMI \cite{EfficientSAM_2024_CVPR}, which also follow the mask-reconstruction paradigm \cite{pan2023towards, DMAE_2023_CVPR}, MG-UKD achieves improvements of 0.6, 1.0, and 0.7 AP over MGD, and 1.2, 0.4, and 0.8 AP over SAMI on AP$^s_S$, AP$^s_M$, and AP$^s_L$, respectively.
These gains can be attributed to the use of a graph attention network (GAT) [50] in MG-UKD, which enables the student model to reconstruct masked patches by leveraging visually similar patches in underwater images. This design helps the student better exploit contextual similarity during the distillation process, leading to improved segmentation performance in challenging underwater environments.

\begin{table}[t]
    \caption{Comparison of MG-UKD with other knowledge distillation methods on the UIIS10K dataset, where w/o KD denotes the absence of knowledge distillation.}
    \vspace{-0mm}
    \centering
    \renewcommand{\arraystretch}{1.4}
    \begin{tabular}{c|cccccc}
    \hline\hline
    Method & $\mathrm{mAP^{s}}$ & $\mathrm{AP^{s}_{50}}$ & $\mathrm{AP^{s}_{75}}$ &  $\mathrm{AP^{s}_{S}}$ & $\mathrm{AP^{s}_{M}}$ & $\mathrm{AP^{s}_{L}}$\\
    \hline
    OFD  \cite{OFD_2019_ICCV} & 38.2 & 54.2 & 41.3 & 7.1 & 28.6 & 49.7\\
    MGD \cite{MGD_2022_ECCV} & 37.8 & 53.9 & 41.1 & 7.1 & 29.0 & 49.3\\
    SAMI \cite{EfficientSAM_2024_CVPR} & 38.1 & 53.9 & 41.2 & 6.5 & 29.6 & 49.2\\
    \rowcolor[RGB]{240,250,239}MG-UKD & 38.7 & 54.3 & 42.1 & 7.7 & 30.0 & 50.0\\
    \hline\hline
    \end{tabular}
    \label{tab:as_mgukd}
    \vspace{-0mm}
\end{table}

\begin{table}[t]
    \caption{Ablation Study on EUPG, w/o CA indicates that the channel attention layer in EUPG is not used.}
    \vspace{-0mm}
    \centering
    \renewcommand{\arraystretch}{1.4}
    \begin{tabular}{c|ccc|ccc}
    \hline\hline
    Method & $\mathrm{mAP^{b}}$ & $\mathrm{AP^{b}_{50}}$ & $\mathrm{AP^{b}_{75}}$ & $\mathrm{mAP^{s}}$ & $\mathrm{AP^{s}_{50}}$ & $\mathrm{AP^{s}_{75}}$  \\
    \hline
    \makebox[1.5cm]{w/o EUPG}  & 39.6 & 53.4 & 43.1 & 36.6 & 51.3 & 40.2 \\
    w/o CA  & 40.8 & 55.6 & 44.9 & 38.2 & 53.9 & 41.8 \\
    \rowcolor[RGB]{240,250,239}EUPG & 41.4 & 56.3 & 45.7 & 38.7 & 54.3 & 42.1 \\
    \hline\hline
    \end{tabular}
    \label{tab:as_eupg}
    \vspace{-2mm}
\end{table}

\noindent\textbf{End-to-End Underwater Prompt Generator}.
To verify the effectiveness of EUPG within the model, we conduct ablation studies by removing it. All experiments are conducted on the UIIS10K dataset, and models are distilled using MG-UKD for 24 epochs.
When EUPG is removed, the model first performs object detection using a ViTDet-style detector \cite{ViTDet_2022_ECCV}~and then passes the predicted bounding boxes into the SAM prompt encoder and mask decoder for instance segmentation. As shown in Table \ref{tab:as_eupg}, this modification leads to a performance drop of 1.8 AP in $ \mathrm{mAP^b} $ and 2.1 AP in $ \mathrm{mAP^s} $.
This degradation can be attributed to the fact that the model needs to first predict the bounding box from the image features and then recode it into prompt features. This repetitive transformation between feature and prompt space introduces unnecessary information loss, and leading to performance degradation.
We also evaluate the contribution of the Channel Attention (CA) layer in the EUPG module by removing it. The results show performance drops of 0.6 AP in $ \mathrm{mAP^b} $ and 0.5 AP in $ \mathrm{mAP^s} $ without CA layer. This suggests that the CA layer enhances the model’s ability to suppress underwater color distortion by dynamically weighting channels, thus improving the feature extraction capability of the model.

\subsection{Conclusion}

In this work, we present \datasetName, the largest underwater instance segmentation dataset to date, containing 10,048 images with pixel-level annotations across 10 object categories. This dataset provides a valuable benchmark for the research community and promotes the development of underwater visual perception algorithms.
We also propose \methodName, a novel end-to-end underwater instance segmentation framework. It incorporates MG-UKD, a mask-based knowledge distillation strategy that transfers knowledge from large ViT-Huge SAM models to lightweight ViT-Small models, enhancing segmentation accuracy while reducing computational overhead. Additionally, we introduce EUPG, an underwater prompt generation module that replaces external detectors and prompt encoder by directly generating prompts with spatial and contextual cues, enabling fully end-to-end training and inference.
Extensive experiments conducted on UIIS10K, UIIS, and USIS10K datasets confirm the effectiveness and generalization capability of our method under challenging underwater conditions.
In future work, we plan to expand both the dataset and the model to more complex and dynamic scenarios, such as underwater video instance segmentation and real-time applications in autonomous underwater vehicles.

{
    \small
    \bibliographystyle{IEEEtran}
    \bibliography{UIIS10K}
}


 





\end{document}